\documentclass[11pt,a4paper]{article}
\usepackage{times,latexsym}
\usepackage{url}
\usepackage[T1]{fontenc}
\usepackage{graphicx}
\usepackage{amsmath,amssymb}
\usepackage{textcomp}
\usepackage{subcaption}
\usepackage{multirow}
\usepackage{booktabs}
\usepackage{enumitem}
\usepackage{csquotes}
\usepackage[dvipsnames]{xcolor}
\usepackage{algorithm}
\usepackage{algorithmicx}
\usepackage{algpseudocode}
\usepackage[acceptedWithA]{tacl2021v1}

\usepackage{xspace,mfirstuc,tabulary}

\newif\iftaclinstructions
\taclinstructionsfalse 
\iftaclinstructions
\renewcommand{\confidential}{}
\renewcommand{\anonsubtext}{(No author info supplied here, for consistency with
TACL-submission anonymization requirements)}
\newcommand{\instr}
\fi

\iftaclpubformat 

\else

\fi

\usepackage{mathtools}

\DeclareMathOperator*{\argmax}{arg\,max}

\newcommand{\usr}{\textrm{usr}}
\newcommand{\sys}{\textrm{sys}}

\definecolor{darkred}{RGB}{204,1,0}

\title{All Learning Has an Emotional Basis, So Does Task-Oriented Dialogue}

\author{Shutong Feng, Hsien-chin Lin, Nurul Lubis, Carel van Niekerk, Michael Heck,\\{\bf Benjamin Ruppik, Renato Vukovic, Milica Ga\v{s}i\'{c}} \\
  Heinrich Heine University Düsseldorf, Germany\\
  \texttt{\small{\{fengs,linh,lubis,niekerk,heckmi,ruppik,revuk100,gasic\}@hhu.de}} \\}

\date{}

\begin{document}
\maketitle
\begin{abstract}
  Task-oriented dialogue (ToD) systems aim to help users accomplish goals through natural language interaction. 
  Beyond task success, effective ToD systems must also maintain positive emotional interaction and accurately convey information in inherently noisy and ambiguous conversational environments. 
  Recent advances in large language models (LLMs) have substantially improved conversational fluency and contextual understanding.
  However, although emotion has been incorporated into existing ToD systems, its integration is typically confined to limited roles, restricting its influence on overall dialogue behaviour and user experience.
  To address this, we propose a novel end-to-end framework that combines fully lexicalised representations with an LLM backbone for improved semantic accuracy and employs online reinforcement learning with short-term emotional and long-term task-success rewards for joint optimisation of emotional interaction and task success. 
  We further develop a challenging and realistic simulation environment that enables systematic evaluation across diverse ToD set-ups.
  Experiments show that incorporating coarse affective signals improves task success and sentiment rating in aggregate.
  Furthermore, integrating emotional interaction throughout the dialogue system using our proposed framework results in more pronounced improvements in both task success and user emotional experience.\footnote{Code can be found at https://gitlab.cs.uni-duesseldorf.de/general/dsml/luster-public.}
\end{abstract}
\section{Introduction}
\label{sec:introduction}

Conversational artificial intelligence (AI) enables machines to engage in natural, human-like interactions using spoken or written language.
The rapid advancement of large language models (LLMs) has significantly enhanced conversational AI~\citep{llama3,qwen35}. 
Trained on vast amounts of text data, LLMs can generate fluent, context-aware responses and follow complex instructions, significantly improving the capabilities of chat applications~\citep{gpt4,deepseek}.

\begin{figure}
    \centering
    \includegraphics[width=0.495\textwidth]{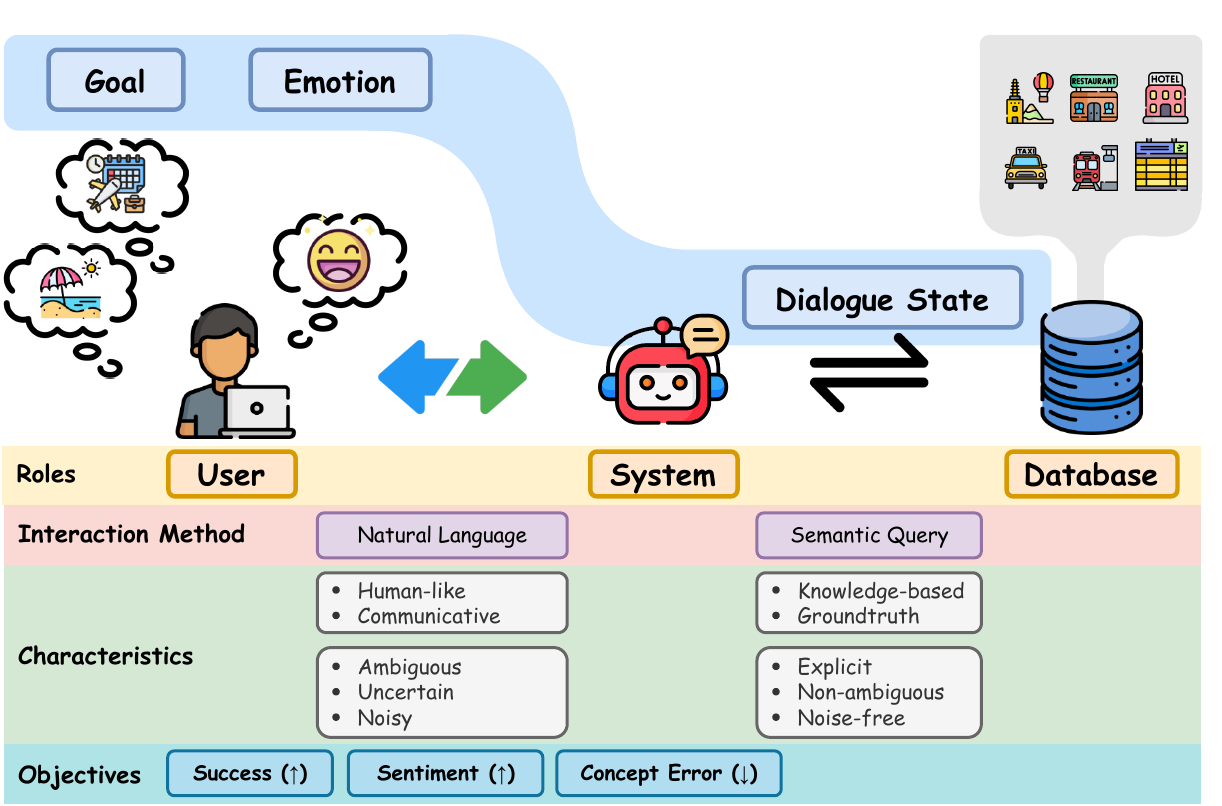}
    \caption{Goal-driven Human-machine Interaction}
    \label{fig:overview}
\end{figure}

Among these applications, task-oriented dialogue (ToD) systems aim to help users accomplish specific goals through multi-turn natural language interaction (\citealt{speech-language-processing}, see Figure~\ref{fig:overview}).
Traditionally, ToD optimisation has focused primarily on task success.
However, effective interactions also require maintaining positive emotional experiences and conveying information accurately in inherently noisy and ambiguous conversational settings~\citep{affective-computing,emowoz}.
High-performing ToD systems must therefore balance task success, emotional interaction, and semantic correctness.

In recent years, there has been growing interest in incorporating emotional intelligence into ToD systems.
However, despite this progress, emotion is typically integrated in a limited scope.
In modular systems, it is often used as an additional reward signal for policy optimisation; in end-to-end models, it appears as an auxiliary supervision objective; and in agentic systems, it is implicitly addressed through general-purpose preference alignment.
Consequently, emotion influences behaviour locally rather than shaping system-wide dialogue behaviour.
Improvements in module performance or model scale do not necessarily translate into better emotional interaction.

To investigate the impact of emotion, we use ``emotion'' in a deliberately well-defined and operational scope. 
Our question is therefore narrower and better controlled: can these structured affective signals be integrated into end-to-end ToD optimisation and used alongside task success to shape system behaviour?

In this work, we treat emotional interaction as a first-class objective in ToD and investigate how its complete integration into end-to-end system optimisation can shape dialogue behaviour in useful ways, especially in repair, clarification, and user-experience-sensitive situations.

Our contributions are threefold:

    
\begin{itemize}
    \item We propose LUSTER, an end-to-end task-oriented dialogue framework that integrates coarse affective interaction signals into system-level optimisation. 
    LUSTER combines fully lexicalised generation with an LLM backbone and online RL using short-term affective and long-term task-success rewards.
    
    \item We evaluate LUSTER across multiple LLM backbones, reward configurations, and human and automatic evaluations. 
    Results show that affect-aware optimisation improves task success and sentiment rating in aggregate, while its effect on semantic accuracy is mixed across configurations.

    \item We provide a behavioural analysis of human evaluation logs showing that the main differences induced by affect-aware optimisation arise in repair, clarification, and affective closure behaviour. 
\end{itemize}

\section{Related Work}

\begin{figure*}[t]
    \centering
    \begin{subfigure}{0.495\textwidth}
        \centering
        \includegraphics[width=\linewidth]{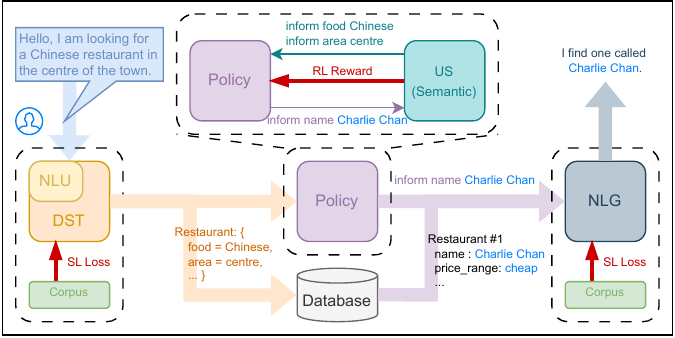}
        \caption{Typical Modular system with independently optimised components.}
        \label{fig:modular-1}
    \end{subfigure}
    \hfill
    \begin{subfigure}{0.495\textwidth}
        \centering
        \includegraphics[width=\linewidth]{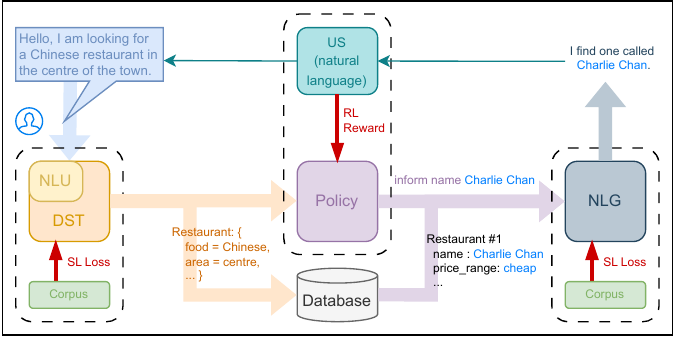}
        \caption{Modular system trained via natural language interactions.}
        \label{fig:modular-2}
    \end{subfigure}
    
    
    \begin{subfigure}{0.495\textwidth}
        \centering
        \includegraphics[width=\linewidth]{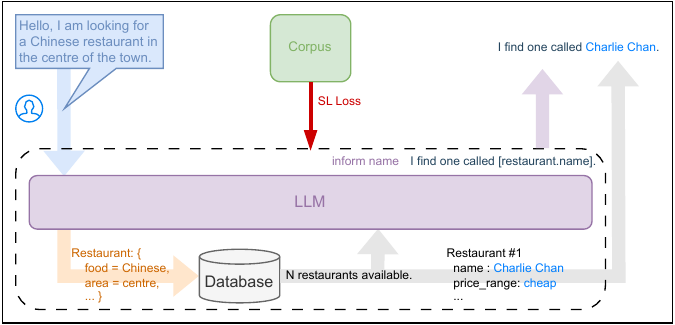}
        \caption{Typical LM-based end-to-end system with SL objective on the corpus.}
        \label{fig:e2e-1}
    \end{subfigure}
    \hfill
    \begin{subfigure}{0.495\textwidth}
        \centering
        \includegraphics[width=\linewidth]{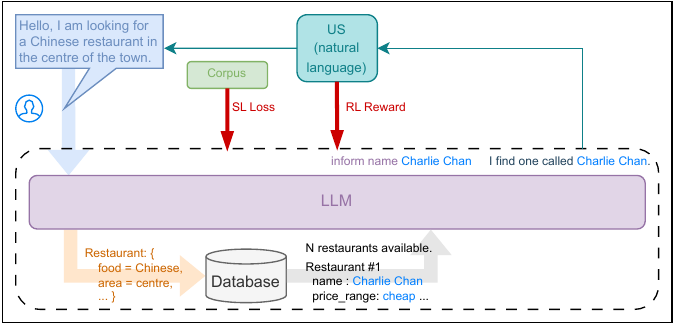}
        \caption{LM-based end-to-end system with SL and RL objectives (ours).}
        \label{fig:e2e-2}
    \end{subfigure}
    \caption{\small ToD system training paradigms. \textcolor{darkred}{Red arrows} indicate learning signals for components in dotted boxes.}
    \label{fig:system_comparison}
\end{figure*}

\subsection{Modular Systems}
\label{sec:related-work:modular-systems}

Traditional ToD systems adopt a modular pipeline consisting of natural language understanding (NLU), dialogue state tracking (DST), dialogue policy learning, natural language generation (NLG), and database querying supported by an ontology~\citep{young2002talking}.
These components are typically trained independently, with NLU, DST, and NLG optimised via supervised learning (SL) and the dialogue policy trained using reinforcement learning (RL, \citealt{pomdp} through interaction with real~\citep{online-rl-real-user} or simulated users~\citep{gentus}.

To manage the complexity of the action space, policies commonly output delexicalised actions, which are later realised into natural language responses using database values.
While this abstraction simplifies policy learning, it separates decision-making from surface realisation and limits joint optimisation across components.

Emotion integration in modular systems primarily occurs at the policy level.
Prior work introduces emotion signals as additional policy inputs~\citep{emotion-as-state} or auxiliary RL rewards~\citep{emotion-as-reward}, enabling policies to adapt to users' affective states.
However, optimisation remains confined to the policy, while other modules are fixed after independently training.
As a result, emotional signals influence behaviour locally without reshaping system-wide representations, leading to misalignment among modules during full-system interaction.

Recent work, such as EmoLoop~\citep{emoloop}, partially mitigates this limitation by optimising an emotion-aware policy using natural-language interaction (Figure~\ref{fig:modular-2}).
Nevertheless, only policy parameters are updated, leaving upstream and downstream components unchanged. 
The long-standing problem of error propagation across modular pipelines remains unresolved.

\subsection{End-to-end Systems}
\label{sec:related-work:e2e-systems}

To address limitations of modular pipelines, end-to-end ToD systems unify dialogue modelling within a single learnable architecture.
Early work builds fully trainable neural models~\citep{e2e-trainable-network}, while recent approaches leverage decoder-only language models to autoregressively map dialogue history to belief states, dialogue actions, and system responses~\citep{simpletod,soloist} (Figure~\ref{fig:e2e-1}).

Despite this unified formulation, many end-to-end systems retain design choices inherited from modular pipelines, particularly the use of delexicalised placeholders that are later replaced with database values.
While this abstraction facilitates SL and evaluation, it separates semantic grounding from response generation and may cause fragility during interactive deployment~\citep{emoloop}.

End-to-end ToD systems are predominantly trained via SL, with some works incorporating RL to improve performance, for example, through behaviour cloning with critic-based filtering~\citep{gpt-critic} or latent action representations~\citep{lava}.
However, these strategies primarily target task success and response quality.

Emotion has been incorporated through auxiliary objectives such as emotion recognition ~\citep{simplellama}, enabling models to condition responses on inferred user affect.
While such multi-task formulations improve performance, emotional interaction remains a supplementary supervision signal rather than a system-level optimisation objective, limiting its influence on overall dialogue behaviour.

\subsection{Agentic Systems}
The emergence of general-purpose LLMs has enabled agentic ToD systems that perform dialogue reasoning through prompting rather than task-specific training.
These systems adapt to new tasks using instructions or demonstrations in zero- or few-shot settings, often employing techniques such as chain-of-thought prompting or self-refinement to improve reasoning consistency~\citep{cot,react}.

Such capabilities have been applied to system components (slot filling~\citep{chatgpt-sl}, dialogue state tracking~\citep{chatgpt-dst}, emotion recognition~\citep{chatgpt-erc}, ontology construction~\citet{ontology-construction-cot}) and fully prompt-driven ToD agents~\citep{fnctod}.
While highly flexible, agentic systems remain sensitive to prompt design, exhibit variability across interactions, and lack the robustness achieved by task-specific optimisation~\citep{llms-all-you-need-tod}.

Emotion in agentic systems is incorporated primarily through implicit alignment or prompting.
RL from human feedback (RLHF) encourages generally polite or empathetic behaviour~\citep{instructgpt}.
However, such preferences constitute multiple reward factors and are not explicitly aligned with the core objectives of ToDs, namely short-term user sentiment and long-term task success.
Alternatively, emotional behaviour can be specified through prompt instructions, making emotion a behavioural guideline imposed at inference time rather than a learnable optimisation objective.
Hence, emotional interaction is not explicitly optimised during dialogue learning in existing agentic ToD systems.

Prior dialogue systems have used categorical affect or sentiment labels for emotion recognition, response style control, and policy optimisation. 
Across modular, end-to-end, and agentic paradigms, however, such signals typically influence behaviour locally rather than serving as system-level optimisation objectives. 
This motivates our development of LUSTER, a unified end-to-end ToD framework that incorporates coarse affective signals directly into optimisation.
Our contribution is therefore not the use of categorical affect itself, but a framework for treating such affective signals as optimisation targets alongside task success, and an empirical study of how this design interacts with task completion, post-dialogue user sentiment, semantic expression, and observable dialogue behaviour.
This distinction is important because affective style alone does not guarantee task success, while task success alone does not guarantee a positive user experience.

\begin{figure*}[ht]
    \centering
    \includegraphics[width=\textwidth]{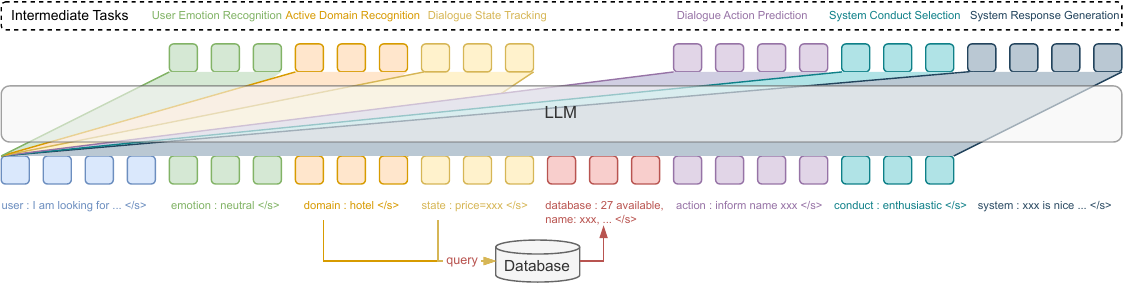}
    \caption{LUSTER model architecture and supervised training objectives.}
    \label{fig:llmtod}
\end{figure*}

\section{LUSTER: \textbf{L}LM-based \textbf{U}nified \textbf{S}ystem for \textbf{T}ask-oriented dialogue with \textbf{E}nd-to-end \textbf{R}einforcement Learning}
\label{sec:luster}
LUSTER integrates emotional interaction into both inference and optimisation, enabling dialogue behaviour to be jointly shaped by task success and affective outcomes.
The framework adopts an end-to-end architecture built on a pretrained LLM backbone, operates on fully lexicalised sequences, and employs online RL with short-term emotional and long-term task-success rewards.

Section~\ref{sec:inference} describes the sequential inference process, while Sections~\ref{sec:lexicalised-sequence} and~\ref{sec:mixed-training-objectives} present the key design choices distinguishing LUSTER from prior LM-based end-to-end ToD systems.
Section~\ref{sec:user-system-emotion} describes the scope of emotion used in this work.

\subsection{Inference with LUSTER}
\label{sec:inference}
LUSTER models ToD as a sequence of structured generation stages in which intermediate semantic and affective representations are explicitly generated and reused to guide subsequent decisions (Figure~\ref{fig:llmtod}).
Beyond traditional ToD subtasks (DST, policy learning, NLG), the framework incorporates user emotion recognition and system conduct selection as intermediate decisions, allowing affective information to influence downstream reasoning and response generation.

Unlike prior LM-based end-to-end systems that separate generation stages using special tokens (e.g., \texttt{<|state|>}), LUSTER adopts a lightweight prefix-based format inspired by~\citet{prefix-tuning}, where each task begins with a natural-language cue (e.g., ``state :'') and terminates with a standard end-of-sequence token.
This design improves compatibility with pretrained LLM tokenisers and stabilises training under parameter-efficient fine-tuning.

At each dialogue turn, LUSTER generates intermediate representations in the following fixed sequence.

\paragraph{Step 0: Dialogue History Encoding} 
At the $t$-th turn, LUSTER receives a dialogue history $\mathcal{H}_t$, constructed as an alternating sequence of user and system utterances, concluding with the most recent user input:
\begin{equation}
\footnotesize
        \mathcal{H}_t 
        \coloneqq
        \{\text{user :} \ u_{\usr,1}, \text{system :} \ u_{\sys,1}, \ldots, \text{user :} \ u_{\usr,t}\}
\end{equation}

\paragraph{Step 1: User Emotion Recognition}
The model first predicts the user's emotional state $e_t$ by selecting the token from a discrete set of emotions $E$ corresponding to the highest logit score:
\begin{equation}
\footnotesize
        e_t = {\argmax}_{e\in E} \ \text{LLM}(\mathcal{H}_t, \text{{``emotion :''}}).\text{logit}(e)
    \label{eqn:emotion}
\end{equation}

\paragraph{Step 2: Active Domain Recognition}
Next, the model identifies the active domain $d_t$ from the predefined set of domains $D$:
\begin{equation}
\footnotesize
    d_t = {\argmax}_{d\in D} \ \text{LLM}(\mathcal{H}_t, e_t, \text{{``domain :''}}).\text{logit}(d)
    \label{eqn:domain}
  \end{equation}
  
\paragraph{Step 3: Dialogue State Tracking}
The dialogue state $ds_t$ is generated for the selected domain to maintain domain relevance and reduce unnecessary context expansion.
All $m$ ontology slots are filled with generated values, candidate selections, or ``unknown'' when absent:

\begin{equation}
\footnotesize
    ds_t \coloneqq \{\text{slot}_1: \text{value}_1 \text{;} \ ... \text{;} \ \text{slot}_m: \text{value}_m \}
\end{equation}
\begin{equation}
    \footnotesize
    ds_t = \text{LLM}(\mathcal{H}_t, e_t, d_t,  \text{``state :''})
    \label{eqn:state}
\end{equation}

\paragraph{Step 4: Database Query}
Matching entries $ent$ are retrieved from the database based on the dialogue state. 
The database description $db_t$ includes the number of matching entities and all \underline{attr}ibute-\underline{val}ue pairs of the first matched entity:

\begin{equation}
\footnotesize
    \text{ent} = \text{database.query}(d_t, ds_t) 
\end{equation}
\begin{equation}
\footnotesize
    db_t \coloneqq \{\text{len(ent) found : } \text{attr} \ \text{val} \ \forall (\text{attr} , \text{val}) \in \text{ent}[0]\}
\end{equation}

\paragraph{Step 5: Dialogue Action Prediction}
Next, the model generates dialogue actions $da_t$ as intent-slot-value tuples:
\begin{equation}
\footnotesize
    da_t \coloneqq \{\text{intent}_1 \ \text{slot}_1 \ \text{value}_1 ;\ ... \}
\end{equation}
\begin{equation}
\footnotesize
    da_t = \text{LLM}(\mathcal{H}_t, e_t, d_t, ds_t, db_t, \text{``action :''})
\end{equation}

\paragraph{Step 6: System Conduct Selection}
The system's emotional stance $c_t$ is then selected from a set of conducts $C$, determining the affective behaviour in the final response.

\begin{equation}
\footnotesize
\begin{aligned}
    c_t = \argmax_{c \in C} \ \text{LLM}(&\mathcal{H}_t, e_t, d_t, ds_t, db_t,  da_t, \\
    &\text{``conduct :''}).\text{logit}(c)
\end{aligned}
\label{eqn:conduct}
\end{equation}

\paragraph{Step 7: System Response Generation}
The final response $u_{\sys,t}$ is generated conditioned on the full context:
\begin{equation}
\footnotesize
    u_{\sys,t} = \text{LLM}(\mathcal{H}_t, e_t, d_t, ds_t, db_t, da_t, c_t, \text{\footnotesize{``system :''}})
\end{equation}

The response is jointly conditioned on user emotion, semantic state, database grounding, dialogue actions, and affective conduct, where emotion influences both reasoning and generation.

\subsection{Fully Lexicalised Representation}
\label{sec:lexicalised-sequence}
LUSTER adopts a fully lexicalised representation in which both intermediate reasoning and final responses are expressed directly in natural language.
Unlike modular and prior end-to-end systems that rely on delexicalised placeholders, LUSTER preserves concrete slot values throughout inference, leveraging the world knowledge and generative capabilities of pretrained LLMs (Figure~\ref{fig:e2e-2}).
Specifically, (1) \textbf{dialogue actions} include explicit values alongside intents and slots, (2) \textbf{responses} are generated directly without placeholder substitution, and (3) \textbf{database results} are represented as structured, value-rich entity descriptions rather than abstract summaries (e.g., ``3 options found'').

This design removes the need for post-hoc lexicalisation and avoids modifying the LLM vocabulary or resizing embedding and output layers, reducing the number of trainable parameters even under parameter-efficient fine-tuning.
Such efficiency is critical for enabling online RL in interactive environments, which would otherwise be computationally prohibitive for large LLMs.

Beyond efficiency, lexicalised representations improve modularity and forward compatibility, allowing LUSTER to adopt newer LLM backbones without redesigning symbolic interfaces or output vocabularies.

\subsection{User and System Affective States}
\label{sec:user-system-emotion}
LUSTER uses affective annotations derived from EmoWOZ \citep{emowoz} rather than introducing a new emotion taxonomy. 
Both the user simulator and the system-side user-emotion recognition component use the seven user emotion categories defined in EmoWOZ, namely \emph{Neutral}, \emph{Fearful}, \emph{Dissatisfied}, \emph{Apologetic}, \emph{Abusive}, \emph{Excited}, and \emph{Satisfied}. 
In addition, the system's emotional conduct follows the five conduct categories introduced in EmoWOZ 2.0 \citep{emoloop}: \emph{Neutral}, \emph{Compassionate}, \emph{Apologetic}, \emph{Enthusiastic}, and \emph{Appreciative}. 
These labels are predicted and generated as part of the dialogue trajectory, allowing LUSTER to represent not only what the user is trying to achieve, but also the affective state expressed by the user and the emotional conduct selected by the system.


\subsection{Training Objectives}
\label{sec:mixed-training-objectives}

LUSTER is trained in two stages: SL provides behavioural grounding from human demonstrations, followed by RL that jointly optimises emotional experience and task success through online interaction.

\subsubsection{Supervised Learning}
\label{sec:supervised-training}

We first train the model to imitate successful dialogue demonstrations from human operators in the EmoWOZ corpus~\citep{emowoz}. 
Each training example is formatted from a dialogue turn into a fully lexicalised sequence $x=\{\mathcal{H}_t, e_t, d_t, ds_t, db_t, da_t, c_t, u_{\sys,t}\}$, following the inference structure described in Section~\ref{sec:inference}. Here, $x$ is a token sequence of length $n=|x|$, denoted by $(x_1, x_2, ..., x_n)$.

LUSTER is trained using the standard causal language modelling objective~\citep{gpt2}.
Given a dataset $\mathcal{X}=\{x^1, x^2, ..., x^{|\mathcal{X}|}\}$, we optimise the model parameters $\theta$ to minimise the negative log-likelihood of the tokens in each sequence:
\begin{equation}
\footnotesize
p(x)=\prod_{i=1}^n p(x_i|x_{<i}), \ \ 
    \mathcal{L}_{SL} = -\sum_{j=1}^{|\mathcal{X}|}\sum_{i=1}^{n_j}\log p_{\theta}(x_i^j|x_{<i}^j)
    \label{eqn:loss-sl}
\end{equation}

This stage provides a behavioural prior that enables coherent task-oriented reasoning and affective responses before RL further refines dialogue behaviour through interaction.

\subsubsection{Reinforcement Learning with Emotion Reward}
\label{sec:single-turn-rl}

While SL provides a behavioural foundation, it does not optimise system behaviour for interaction outcomes.
We therefore introduce RL to directly shape responses using emotion feedback.
Following the RLHF formulation, dialogue generation is modelled as a sequential decision-making process in which the LLM policy $\pi_\theta$ produces tokens (actions $a$) conditioned on dialogue context (states $s$), with optimisation guided by emotion-derived rewards.

\paragraph{Episode Formulation}
An episode corresponds to generating the system response at the $t$-th interaction, resulting in the following sequence:

\begin{equation}
\footnotesize
    \{\mathcal{H}_t, e_t, d_t, ds_t, db_t, da_t, c_t, u_{\sys,t}\}
\end{equation}
To limit context length, we set the initial state as:
\begin{equation}
\footnotesize
    s_t^0 = \{\mathcal{H}_t, e_t, d_t, ds_t, db_t\}
    \label{eqn:turn-level-state-1}
\end{equation}
Note that superscripts denote the time step in the generation of a single turn, while the subscript $t$ refers to the dialogue-level time step, which remains fixed. 

Each action token $a_t^l$ is then generated from the policy $\pi_\theta(a_t^l|s_t^l)$. 
We obtain the reward $r_t^{l+1}$, and the state is incrementally updated:

\begin{equation}
\footnotesize
    s_t^{l+1} = \{s_t^0, a_t^{<l}\}
\end{equation}
At the end of the episode, we have $L$ action tokens:
\begin{equation}
\footnotesize
    a_t^{0:L-1} = \{da_t, c_t, u_{\sys,t}\}
    \label{eqn:turn-level-action-all}
\end{equation}


\paragraph{Reward Modelling via Emotion Advantage}
Rather than rewarding absolute emotional outcomes, we optimise relative emotional improvement among alternative responses generated under the same context.
Given system response $u_{\sys,t}$, a user simulator (US) produces the next user utterance and emotion:
\begin{equation}
\footnotesize
    u_{\usr,t+1}, e_{t+1} = \text{US}(u_{\sys,t})
\end{equation}

Multiple trajectories are sampled from the same initial state:
\begin{equation}
\footnotesize
    \tilde{a}_t^{0:L-1} = \{\widetilde{da}_t, \tilde{c}_t, \tilde{u}_{\sys,t}\} \sim \prod_{l=1}^{L} \pi(a_t^l|s_t^l)
    \label{eqn:sequence-sample}
\end{equation}

One of the sampled sequences is then randomly selected to continue the conversation with the US. 
Each sampled response $\tilde{u}_{\sys,t}$ is evaluated by the US for a resulting emotion $\tilde{e}_{t+1}$, which is mapped to a valence-based numeric reward with the mapping function $\mathcal{M}(\cdot)$: $+1$ for positive, $0$ for neutral, and $-1$ for negative. 
Rewards for non-terminal tokens are set as $r^{1:L-1}=0$, and the final reward is normalised across samples:

\begin{equation}
\footnotesize
   \tilde{r}_t^L=\frac{\mathcal{M}(\tilde{e}_{t+1})-\mathbb{E}[\mathcal{M}(\tilde{e}_{t+1})]}{\sigma(\mathcal{M}(\tilde{e}_{t+1}))}
   \label{eqn:emotion-advantage}
\end{equation}

We refer to this signal as the \textbf{emotion advantage}, 
as it captures the relative emotional improvement compared to alternative responses.
In practice, samples with zero variance are discarded.

\paragraph{Policy Optimisation with Emotion Advantage}
We maximise the expected return of the policy $\pi_\theta$ across the sampled trajectories with turn-level discount factor $\gamma_{\text{turn}}$. 
\begin{equation}
\footnotesize
   J(\theta)=\mathbb{E}_{\pi_\theta}\sum_{l=0}^{L-1}\gamma_{\text{turn}}^{l}r_t^{l+1}
\end{equation}

Gradients are estimated using the policy gradient theorem and the advantage function $A_{\pi_\theta}(s_t^l,a_t^l)$~\citep{pg}:
\begin{equation}
\footnotesize
    \nabla_\theta J(\theta) = \mathbb{E}_{\pi_\theta}\sum_{l=0}^{L-1} \nabla_\theta \log \pi_\theta(a_t^l|s_t^l) A_{\pi_\theta}(s_t^l,a_t^l)
    \label{eqn:pg}
\end{equation}

In principle, this objective can be optimised using any actor-critic policy optimisation algorithm. 
In our implementation, we instantiate it with Proximal Policy Optimisation (PPO)~\citep{ppo}; full implementation details are provided in Appendix~\ref{sec:appendix:ppo}.


\subsubsection{Reinforcement Learning with Success Reward}
\label{sec:multi-turn-rl}
While turn-level optimisation encourages the elicitation of positive emotions, task success requires planning across multiple turns.
We introduce a complementary dialogue-level RL objective that captures long-term task performance.

\paragraph{Episode Formulation} An episode corresponds to a full dialogue, consisting of $T$ user-system turns. 
At the $t$-th step in the dialogue, we derive the state and action with a formatter function $\mathcal{F}(\cdot)$ from Equations~\ref{eqn:turn-level-state-1} and~\ref{eqn:turn-level-action-all}: the dialogue-level state $\mathcal{S}_t=\mathcal{F}(s_t^0)$ and dialogue-level action $\mathcal{A}_t=\mathcal{F}(a_t^{0:L-1}$) (see Appendix \ref{sec:appendix:critic} for formatter details). 
This forms a sequence of state-action pairs $\{(\mathcal{S}_0, \mathcal{A}_0), ..., (\mathcal{S}_{T-1}, \mathcal{A}_{T-1})\}$ over the course of the interaction.

\paragraph{Reward Modelling}
The dialogue-level reward is based on task success and interaction efficiency. 
We train two dialogue-level critic networks for our policy $\pi$ parameterised by $\phi$ and $\psi$: state-action value function $Q_\phi^\pi(\mathcal{S}_t,\mathcal{A}_t)$ and state value function $V_\psi^\pi(\mathcal{S}_t)$.
These critics estimate long-term dialogue outcomes, enabling optimisation beyond immediate emotional feedback.
Both critics are based on ALBERT~\citep{albert}, with a linear projection on the \texttt{[CLS]} token embedding to produce scalar value estimates. 
Following the Archer framework~\citep{archer}, they are trained to minimise temporal difference (TD) error and ensure Bellman consistency using online experience $\mathcal{E}=\{\mathcal{S}_t,\mathcal{A}_t,r_{t+1},\mathcal{S}_{t+1}\}_{t=0}^{T-1}$ collected during interactions with a user simulator:
\begin{equation}
\footnotesize
    \mathcal{L}_Q = \mathbb{E}_{s,a,r,s'\sim\mathcal{E}}[(Q_\phi(s,a)-r-\gamma_{\text{dial}}V_{\bar{\psi}}(s'))^2] \label{eqn:obj-q}
\end{equation}
\begin{equation}
\footnotesize
    \mathcal{L}_V = \mathbb{E}_{s\sim \mathcal{E}}\big[
        \mathbb{E}_{a\sim\pi_{\theta}(\cdot|s)}
        [(V_{\psi}(s)-Q_{\bar{\phi}}(s,a))^2]
    \big]\label{eqn:obj-v}
\end{equation}
The delayed target models $Q_{\bar{\phi}}$ and $V_{\bar{\psi}}$ are updated towards their current counterparts with Polyak averaging~\citep{polyak}.
Refer Appendix~\ref{sec:appendix:critic} for critic training details.

Given the trained critics, we compute the advantage estimate at each $t$-th turn using:
\begin{equation}
\footnotesize
    A_t=Q_\phi^\pi(\mathcal{S}_t,\mathcal{A}_t) - V_\psi^\pi(\mathcal{S}_t)
   \label{eqn:success-advantage}
\end{equation}

This value measures how much better a particular action is compared to the expected return of the current state. 
We use this signal in the same PPO optimisation framework as described in Section~\ref{sec:single-turn-rl}, replacing the emotion advantage with this task-oriented multi-turn advantage.

\paragraph{Combined Reward}
The combined reward balances short-term positive emotion elicitation with long-term task effectiveness.
The single-turn (Equation \ref{eqn:emotion-advantage}) and multi-turn reward (Equation \ref{eqn:success-advantage}) signals are combined for selected sequences from Equation \ref{eqn:sequence-sample} using a weighted sum with mixing parameter $\rho$:
\begin{equation}
\footnotesize
    \mathcal{R}_{\text{combine},t}=\rho A_t+r_t^L \label{eqn:combined-reward}
\end{equation}

\paragraph{Off-policy Experience Replay}
Because simulated interaction is computationally expensive, we maintain an experience buffer and reuse past trajectories via importance sampling (IS, \citealt{importance-sampling}).

A constant buffer size is enforced using reservoir sampling~\citep{reservoir-sampling}. The IS weight is defined as:
\begin{equation}
\footnotesize
    \alpha_t=\frac{\pi_{\theta_\text{new}}(\mathcal{A}_t|\mathcal{S}_t)}{\pi_{\theta_\text{old}}(\mathcal{A}_t|\mathcal{S}_t)}
\end{equation}

In practice, computing IS weights over long sequences can be numerically unstable due to the sharp trajectory distribution of LLMs. 
This often leads to high variance in IS weight estimates, even when the behaviour and target policies only diverge slightly. 
To mitigate this, we compute IS weights only over intents and slots of dialogue actions $da_{t,\text{intent+slot}}$, and clip them to $[1-\eta, 1+\eta]$:
\begin{equation}
\footnotesize
    \hat{\alpha}_t=\text{clip}\Big(\frac{\pi_{\theta_\text{new}}(da_{t,\text{intent+slot}}|\mathcal{S}_t)}{\pi_{\theta_\text{old}}(da_{t,\text{intent+slot}}|\mathcal{S}_t)}, 1-\eta,1+\eta\Big)
    \label{eqn:is-weight}
\end{equation}
When deploying experience replay, we multiply the corresponding reward for past experiences (Equation \ref{eqn:success-advantage} or \ref{eqn:combined-reward}) with IS weight $\hat{\alpha}_t$.

Through this two-level optimisation scheme, LUSTER jointly learns to produce emotionally appropriate responses at the turn level while improving task success over entire dialogues.

\section{Experimental Set-up}

\subsection{ToD Systems}
\label{sec:tod-systems}

We evaluate emotion as an independent design dimension across agentic, modular, and end-to-end ToD paradigms.
For each system, we compare a non-emotional variant with an emotion-aware counterpart whenever applicable, allowing us to isolate the effect of emotion integration while keeping the architecture and training unchanged.
We consider the following systems.

\paragraph{\textbf{FnCTOD}}~\citep{fnctod}
is an \emph{agentic system} that performs prompt-based zero-shot generation for DST, database querying, and response generation.
We follow the original prompts and use GPT-4o-mini (2025-04-16) as the backbone.
Its emotional variant augments the prompt with instructions to consider user emotions and generate emotionally appropriate responses.

\paragraph{\textbf{Modular-base}}
is a modular pipeline consisting of SetSUMBT DST~\citep{setsumbt}, DDPT policy~\citep{ddpt}, and SEC-BART NLG~\citep{emoloop}.
The policy is trained via RL under \emph{semantic} interaction with EmoUS~\citep{emous}, using task-success rewards (Figure~\ref{fig:modular-1}).
\textbf{Modular-emotion} extends this system by adding user emotion as a policy input via an emotion recogniser~\citep{ertod}, producing system conduct decisions and receiving turn-level sentiment rewards.

\paragraph{\textbf{SimpleLoop}}~\citep{emoloop}
uses the same modular architecture as Modular-base but performs RL optimisation within a natural-language interaction loop using langEmoUS~\citep{emoloop} (Figure~\ref{fig:modular-2}).
Its emotional counterpart, \textbf{EmoLoop}, additionally incorporates user emotion input, system conduct output, and sentiment-based rewards.

\paragraph{\textbf{SimpleLLAMA}}~\citep{simplellama}
is an \emph{end-to-end} system that autoregressively generates dialogue state, dialogue action, and response using a 7B {Llama-2-chat} backbone~\citep{llama2}.
It produces delexicalised actions and responses and is trained with SL.
\textbf{EmoLLAMA} extends this model with two additional intermediate tasks: user emotion recognition and system conduct selection.

\paragraph{\textbf{LUSTER}}
is our proposed \emph{end-to-end} framework described in Section~\ref{sec:luster}, trained via RL through natural-language interaction with langEmoUS (Figure~\ref{fig:e2e-2}). 
Its non-emotional counterpart, \textbf{LUSTER-base}, removes emotion-related tasks and RL.
Unless otherwise specified, LUSTER uses {Llama-3.2-3b-instruct}~\citep{llama3} as the backbone.
Additional backbone experiments are reported in Section~\ref{sec:backbones}.

\subsection{Supervised Learning}
All systems and sub-modules are trained on EmoWOZ, an extension of MultiWOZ \citep{multiwoz} annotated with user emotion and system emotional-conduct labels. 
EmoWOZ contains 10,438 dialogues in total, following the standard MultiWOZ split: 8,438 dialogues for training, 1,000 for validation, and 1,000 for testing. 
The training split is used exclusively for SL, while the validation and test splits are held out from SL training.
We do not perform a data-scaling ablation in this work.
In our preliminary experiments, stronger SL initialisation consistently led to better subsequent RL performance, while weak initialisation from limited SL data was difficult to recover through RL alone.
Systematically characterising this dependency remains an important direction for future work.

\subsection{Reinforcement Learning and Interactive Evaluation}

The RL of LUSTER and the subsequent interactive evaluation are conducted using ConvLab-3~\citep{convlab-3}.
We use the natural-language user simulator langEmoUS together with T5NLU~\citep{convlab-3} for dialogue understanding, creating a realistic noisy interaction environment.

This set-up is substantially more demanding than standard policy optimisation and evaluation settings with user simulators.
In conventional RL set-ups, interaction typically occurs at the semantic level~\citep{ddpt}.
Even in natural-language settings, user simulators often still receive semantic actions directly from the policy~\citep{emoloop}, so user goals can be updated unambiguously.
By contrast, in real interaction, users only observe the system's natural language response and do not have access to the underlying semantic action.
As a result, successful communication depends entirely on whether the intended semantics are correctly conveyed through natural language.

To better reflect natural-language interaction, we do not pass semantic actions directly to the simulated user for goal updates or success evaluation. 
Instead, the user side relies on the NLU module to parse the system response and recover the underlying semantics for internal state updates. 
This introduces an additional layer of uncertainty and noise, making evaluation more sensitive to errors in surface realisation. 

For each system, we simulate $3000$ evaluation dialogues. 
These simulated dialogues are separate from the EmoWOZ SL split: they are generated during RL and evaluation, and are not used as supervised training data.
Implementation details, hyperparameters, training configurations, and computational resources are provided in Appendices~\ref{appendix:sec:compute} and~\ref{sec:appendix:hyperparameters}.

The challenging set-up partly explains the substantially lower performance of baseline systems in Table \ref{tab:results} that were originally developed or evaluated under cleaner semantic-interaction settings. 
At the same time, LUSTER's stronger performance under the same protocol highlights the robustness of its fully lexicalised end-to-end optimisation.

\subsection{Evaluation Metrics}
We evaluate each system along three complementary dimensions: task completion, affective interaction, and semantic faithfulness.

\paragraph{Task success.}
Task success measures whether the dialogue fulfils the user's goal. 
The user simulator parses the semantic information expressed by the system and compares it against its internal goal. 
A dialogue is considered successful if the system provides the requested information, the returned entity satisfies the user's search constraints, and a booking is completed when required. 
Task success is therefore a dialogue-level metric: it evaluates whether the interaction as a whole achieves the user's task, rather than whether each individual system response is semantically perfect.

\paragraph{User sentiment.}
User sentiment measures the affective trajectory of the interaction from the simulator's perspective. 
At each user turn, the simulator estimates the user's emotion and maps it to a scalar score based on valence polarity: $+1$ for positive, $0$ for neutral, and $-1$ for negative. 
The reported user sentiment score is the average of these turn-level sentiment values across the dialogue. 
This metric captures whether the interaction tends to leave the simulated user in a more positive or negative affective state, but it should not be interpreted as a complete measure of human emotion.

\paragraph{Concept error rate.}
Concept error rate measures the semantic faithfulness of the generated system response with respect to the system's internal semantic action. 
It detects missing concepts, where semantic values specified by the action are omitted from the surface response, and hallucinated concepts, where the response mentions semantic values that are not licensed by the action. 
Concept error rate is therefore a response-level diagnostic of semantic expression, rather than a direct measure of dialogue-level task completion.

These metrics are related but not interchangeable. 
A system may achieve low concept error while still failing the task because of policy-level mistakes, invalid database choices, missing booking actions, or failure to provide all requested information. 
Conversely, a system may complete the task while still producing occasional semantic omissions or hallucinations. 
Similarly, user sentiment reflects the affective quality of the interaction, but positive sentiment alone does not guarantee task success. 
We therefore report all three metrics to distinguish task performance, affective user experience, and semantic faithfulness.

\section{Results and Discussion}

\subsection{Model Performance}
\begin{table*}[ht]
\centering
\footnotesize
\setlength\tabcolsep{3.5pt}
\begin{tabular}{ll|rrr|rrr}
\toprule
\multicolumn{1}{c}{\multirow{2}{*}{System}} & \multicolumn{1}{c|}{\multirow{2}{*}{Architecture}} & \multicolumn{3}{c|}{without emotion} & \multicolumn{3}{c}{with emotion} \\
\multicolumn{1}{c}{} & \multicolumn{1}{c|}{} & \multicolumn{1}{c}{Success ($\uparrow$)} & \multicolumn{1}{c}{Sentiment ($\uparrow$)} & \multicolumn{1}{c|}{Error ($\downarrow$)} & \multicolumn{1}{c}{Success ($\uparrow$)} & \multicolumn{1}{c}{Sentiment ($\uparrow$)} & \multicolumn{1}{c}{Error ($\downarrow$)} \\
\midrule
FnCTOD & Agentic & 32.3\% & 0.229 & NA & 32.0\%  {\color{gray}\footnotesize(- 0.3\%)} & 0.230 {\color{gray}\footnotesize(+0.001)} & NA \\
Modular-base & Modular & 31.8\% & 0.335 & 5.35\% & 33.0\% {\color{gray}\footnotesize(+1.2\%)} & 0.345 {\color{Green}\footnotesize(+0.010)} & 5.14\% {\color{Green}\footnotesize(-0.21\%)} \\
SimpleLoop & Modular & 35.2\% & \textbf{0.346} & 4.74\% & 37.2\%  {\color{Green}\footnotesize(+2.0\%)} & \textbf{0.380} {\color{Green}\footnotesize(+0.034)} & 3.36\% {\color{Green}\footnotesize(-1.38\%)} \\
SimpleLLAMA & End-to-end & 33.0\% & 0.214 & 5.11\% & 34.2\% {\color{gray}\footnotesize(+1.2\%)} & 0.250 {\color{Green}\footnotesize(+0.036)} & 3.68\% {\color{Green}\footnotesize(-1.43\%)} \\ \midrule
LUSTER (ours) & End-to-end & \textbf{54.4\%} & 0.327 & \textbf{1.69\%} & \textbf{58.0\%} {\color{Green}\footnotesize(+3.6\%)} & 0.352 {\color{Green}\footnotesize(+0.025)} & \textbf{1.71\%} {\color{gray}\footnotesize(+0.02\%)} \\
\bottomrule
\end{tabular}
\caption{Task \textbf{success}, user \textbf{sentiment}, and concept \textbf{error} rate of all systems. Best scores across all systems are \textbf{bolded}. Numbers in brackets denote the difference from the non-emotional system, and green colour denotes statistically significant improvement (p<0.05). The concept error rate of FnCTOD is not available due to the absence of semantic actions.}
\label{tab:results}
\end{table*}

From Table~\ref{tab:results}, incorporating affective signals generally improves user sentiment for trained systems, with gains in task success and concept error observed in several settings.
However, the effect is not uniform across architectures. 
Modular and end-to-end baselines benefit more clearly from affective modelling, whereas the agentic system shows negligible change. 
We therefore interpret these results as evidence that affective signals can improve emotional interaction quality and, in some systems, support task performance and semantic expression.

The limited improvement observed for FnCTOD may be explained by its response strategy.
From dialogue statistics, we observe that FnCTOD produces substantially shorter interactions and longer responses than other systems.
Rather than progressively refining the dialogue state, FnCTOD tends to provide exhaustive information, often returning complete entity descriptions under loose constraints.
This behaviour leaves fewer turns in which affective signals can shape dialogue adaptation, which may limit the observable impact of emotion-aware modelling.

LUSTER achieves the strongest performance, substantially outperforming all baselines in both task success and concept error rate.
This result demonstrates the competitiveness of the proposed fully lexicalised end-to-end framework under a noisy natural-language interaction protocol. 
However, comparisons with external baselines do not by themselves isolate the effect of affective optimisation, since these systems differ in architecture, representation, backbone, and training strategy.

The high task success and low concept error rate of LUSTER are therefore best interpreted as a benefit of the lexicalised end-to-end design and the representational capacity of the LLM backbone, rather than as a direct consequence of affective modelling alone. 
By reasoning directly over entity values, semantic actions, and database information in lexical form, LUSTER can reduce information loss between action selection and surface realisation. 
The effect of affective optimisation is isolated more directly in the ablation study, where LUSTER is compared against LUSTER-base and intermediate reward configurations under matched architecture and training conditions.


\subsection{Ablation Study}
\label{sec:llama-ablation}
\begin{table}[h]
\footnotesize
\setlength\tabcolsep{3pt}
\begin{tabular}{lrrr}
\toprule
 & \multicolumn{1}{c}{Succ. ($\uparrow$)} & \multicolumn{1}{c}{Sent. ($\uparrow$)} & \multicolumn{1}{c}{Error ($\downarrow$)} \\
 \midrule
Base & 54.4\% & 0.327 & 1.69\% \\
Base + EM & 54.7\% & $\dagger$0.342 & \textbf{$\dagger$1.50\%} \\
Base + EM + RL (Sent.) & 49.0\% & \textbf{$\dagger$0.404} & 2.19\% \\
Base + EM + RL (Succ.) & $\dagger$56.9\% & $\dagger$0.339 & 1.89\% \\
LUSTER & \textbf{$\dagger$58.0\%} & $\dagger$0.352 & 1.71\% \\
\bottomrule
\end{tabular}\caption{Task \textbf{succ}ess, user \textbf{sent}iment, and concept \textbf{error} rate of different LUSTER configurations. \textbf{Base} denotes LUSTER-base, and \textbf{EM} denotes supervised emotion modelling, including user emotion recognition and system emotional conduct selection. \textbf{RL (Sent.)} and \textbf{RL (Succ.)} denote RL with \textbf{sent}iment and task-\textbf{succ}ess rewards, respectively. These two RL variants are evaluated independently on top of Base + EM, while the final \textbf{LUSTER} row combines EM with both RL rewards. Best scores within the ablation are bolded. $\dagger$ denotes statistically significant improvement over LUSTER-base ($p<0.05$).}
\label{tab:ablation}
\end{table}

We conduct an ablation study to analyse the contribution of emotion modelling and RL within LUSTER configurations.

As shown in Table~\ref{tab:ablation}, adding emotion modelling via SL (Base + EM) yields a modest sentiment improvement together with lower concept error.
This suggests that user emotion and system conduct labels can provide useful conditioning signals for response quality and semantic grounding, although the latter RL variants show that this effect is not automatic under all optimisation objectives.

Adding RL with a sentiment reward (Base + EM + RL (Sent.)) substantially improves sentiment, achieving the highest sentiment score among the ablation settings, but reduces task success and increases concept error. 
This suggests that optimising affective reward alone can misalign with task completion: the model may elicit positive simulator sentiment even when the response contains hallucinated or incomplete task information.

Optimising only task success (Base + EM + RL (Succ.)) significantly improves success while slightly increasing sentiment, but concept error remains higher than in the supervised settings. 
This suggests that RL can shift the model away from supervised response patterns, improving dialogue-level goal completion while making surface realisation slightly less reliable.

When sentiment and success rewards are jointly optimised (LUSTER), the model achieves the best overall success rate while maintaining improved sentiment and keeping concept error close to the non-emotional baseline. 
This suggests that the two rewards play complementary roles: sentiment reward encourages responses that elicit more positive user affect, while success reward constrains optimisation toward task completion. 
Balancing these objectives improves task success and affective interaction quality while maintaining comparable semantic reliability.


\subsection{LUSTER with Different LLM Backbones}
\label{sec:backbones}
\begin{figure*}[t]
    \centering

    \begin{subfigure}[t]{0.48\textwidth}
        \centering
        \includegraphics[width=\linewidth]{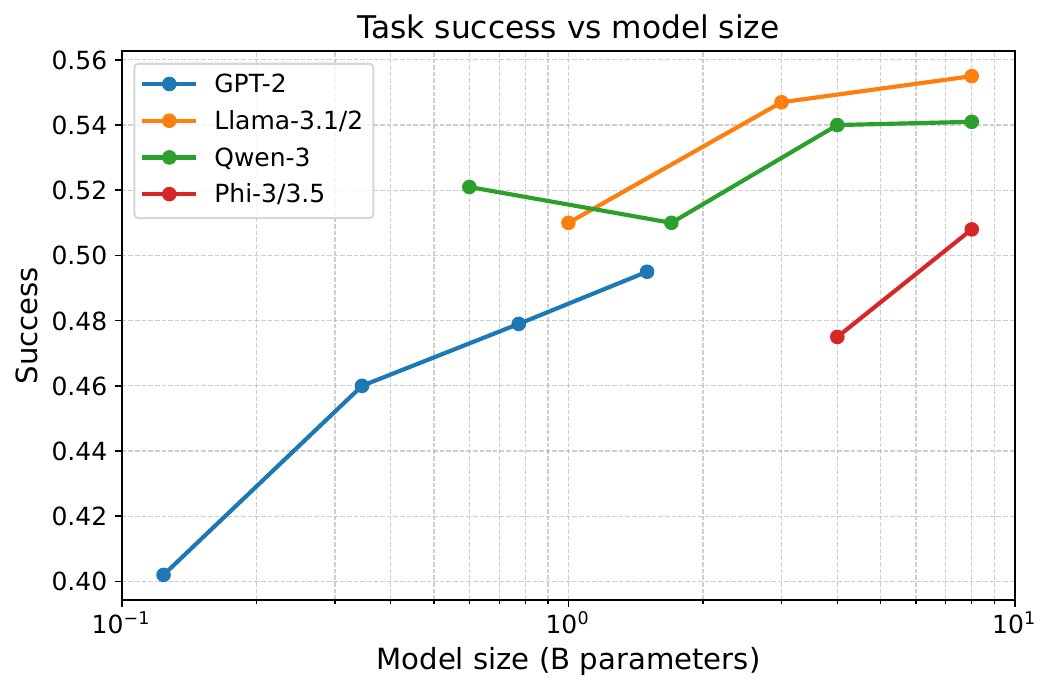}
        \caption{Task success against model size.}
        \label{fig:scale-success}
    \end{subfigure}
    \hfill
    \begin{subfigure}[t]{0.48\textwidth}
        \centering
        \includegraphics[width=\linewidth]{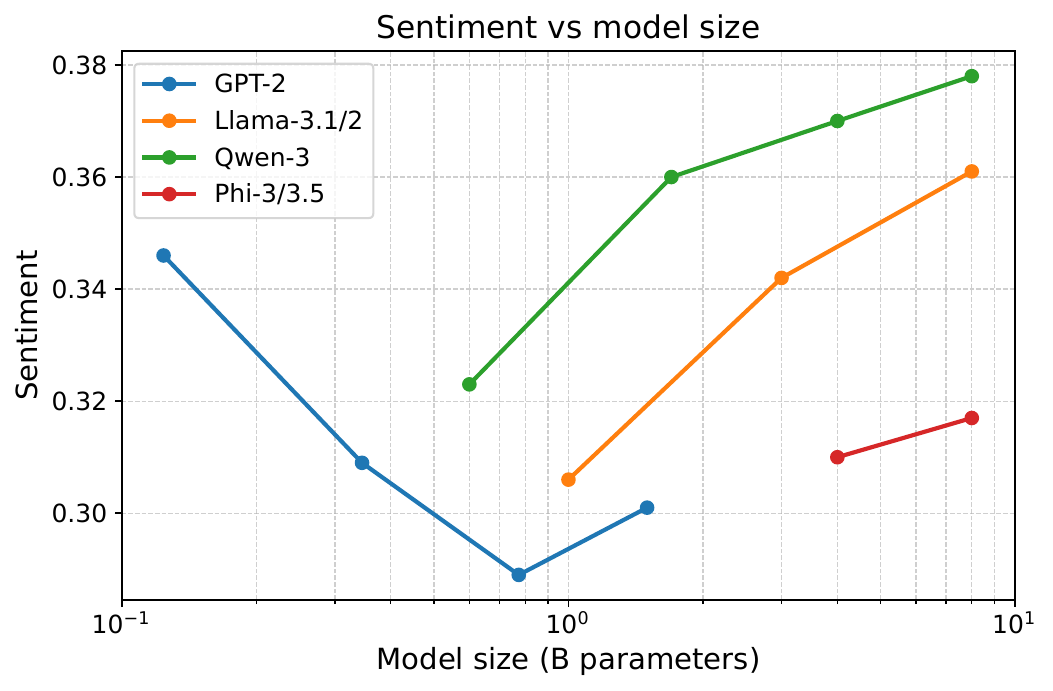}
        \caption{Sentiment score against model size.}
        \label{fig:scale-sentiment}
    \end{subfigure}

    \caption{
    Scaling behaviour across LLM families. Performance is plotted against model size (billions of parameters, log scale). GPT-2 models are GPT-2, GPT-2-medium, GPT-2-large, and GPT-2-xl. Llama-3-1/2 models are Llama-3.2-1B-Instruct, Llama-3.2-3B-Instruct, and Llama-3.1-8B-Instruct. Qwen-3 models are Qwen3-0.6B, Qwen3-1.7B, Qwen3-4B-Instruct-2507, and Qwen3-8B. Phi-3/3.5 models are Phi-3.5-mini-instruct and Phi-3-mini-4k-instruct. 
    }
    \label{fig:scaling}
\end{figure*}

We evaluate several backbone families: Llama-3~\citep{llama3}, Qwen-3~\citep{qwen35}, Phi-3/3.5~\citep{phi-3.5}, and GPT-2~\citep{gpt2}.
Models with at least 3B parameters are fine-tuned using LoRA~\citep{lora}.

\subsubsection{Supervised Learning}
We configure all models to include emotion-modelling intermediate tasks and train them with SL.
We aim to examine scaling behaviour and identify a strong backbone for subsequent RL.
Figure \ref{fig:scale-success} and \ref{fig:scale-sentiment} visualise the trend of each metric against model size.

\textbf{Task success} generally improves with model size across modern LLM families, although gains diminish at larger scales.
This motivates our choice of Llama-3.2-3B for the main experiments, as it provides a balanced trade-off between performance and computational cost.

\textbf{Sentiment} shows a more nuanced trend.
For Llama, Qwen, and Phi, sentiment tends to improve with scale, suggesting that larger models capture affective cues and generate emotionally appropriate responses better.
In contrast, GPT-2 exhibits a non-monotonic trend: sentiment declines at intermediate scales before partially recovering at 1.5B.

One plausible explanation is that smaller GPT-2 models rely heavily on surface-level heuristics or templated responses that may appear emotionally consistent, whereas mid-sized variants lack sufficient representational capacity to robustly model both task semantics and emotional nuance.
Only at larger scales does GPT-2 regain enough capacity to stabilise emotional behaviour, although it still underperforms compared to modern LLMs.

Notably, the divergence between success and sentiment trends highlights that sentiment is not strictly correlated with task success.
This suggests that emotion constitutes an additional optimisation dimension, emphasising the need to explicitly incorporate emotional objectives into dialogue system training rather than assuming they will emerge from the optimisation for task success.

\subsubsection{Reinforcement Learning}
We further investigate whether RL can improve a weaker backbone by applying the same RL training procedure to the worst-performing model in our backbone study, namely {Phi-3.5-mini-instruct} (4B parameters).
This experiment examines whether a model with relatively poor performance after supervised learning can be improved through RL, and whether the effects of emotion modelling and RL remain consistent across backbones.
Accordingly, we repeat the same ablation setting used in Section~\ref{sec:llama-ablation}.

\begin{table}[h]
\footnotesize
\setlength\tabcolsep{3pt}
\begin{tabular}{lrrr}
\toprule
 & \multicolumn{1}{c}{Succ. ($\uparrow$)} & \multicolumn{1}{c}{Sent. ($\uparrow$)} & \multicolumn{1}{c}{Error ($\downarrow$)} \\
 \midrule
Base & 47.2\% & 0.299 & 1.73\% \\
Base + EM & 47.5\% & $\dagger$0.310 & \textbf{$\dagger$1.38\%} \\
Base + EM + RL (Sent.) & $\dagger$49.8\% & \textbf{$\dagger$0.345} & 1.79\% \\
Base + EM + RL (Succ.) & $\dagger$50.8\% & $\dagger$0.324 & 1.51\% \\
LUSTER & \textbf{$\dagger$51.2\%} & $\dagger$0.316 & 1.77\% \\
\bottomrule
\end{tabular}\caption{Ablation of LUSTER configurations using the Phi-3.5-mini-instruct backbone. \textbf{Base}, \textbf{EM}, \textbf{RL (Sent.)}, \textbf{RL (Succ.)}, and \textbf{LUSTER} follow the definitions in Table~\ref{tab:ablation}. Best scores within the ablation are bolded. $\dagger$ denotes statistically significant improvement over LUSTER-base ($p<0.05$).}
\label{tab:phi-ablation}
\end{table}

The overall pattern is generally consistent with the observations in Section~\ref{sec:llama-ablation}.
Adding emotion modelling in SL yields modest improvements in sentiment and reduces concept error.
Using RL with a sentiment reward produces the strongest sentiment performance, but at the cost of higher concept error.
Using RL with a success reward leads to higher task success, a smaller improvement in sentiment, and a slightly worsened concept error rate.
Combining all components gives the best task success while still achieving significantly better sentiment than the Base model.
These observations suggest that the LUSTER framework generalises across LLM backbones.

The final performance of LUSTER with the {Phi-3.5-mini-instruct} backbone remains below the main Llama-based result.
This suggests that SL performance, or more broadly the quality of the initialisation, may signal the potential benefit of subsequent RL.
In practice, this provides a useful and practical criterion for backbone selection before investing additional time and computational resources into RL.

There are, however, two notable differences.
First, RL with a sentiment reward (Base + EM + RL (Sent.)) reduces task success for {Llama-3.2-3b-instruct}, but improves it for {Phi-3.5-mini-instruct}.
We hypothesise that sentiment-only RL is less stable and that different backbones exploit the reward signal in different ways.
This highlights that sentiment reward alone, without stronger regularisation, does not reliably translate into higher task success.

Second, when sentiment and success rewards are jointly optimised (Base + EM + RL (Succ.) $\rightarrow$ LUSTER), both sentiment and concept error improve for {Llama-3.2-3b-instruct}, whereas such improvements are not observed in {Phi-3.5-mini-instruct}.
This suggests that the balance between reward signals is backbone-dependent and requires a more careful weighting.

\subsection{Interactive Evaluation with Real Users}

\subsubsection{Evaluation Set-ups and Human Ratings}
We conduct a human evaluation on Amazon Mechanical Turk using the DialCrowd toolkit~\citep{dialcrowd} to compare LUSTER, LUSTER-base, and EmoLoop.
Participants are presented with randomly generated single- or multi-domain goals specifying entity constraints and required information.
After they interact with each system to complete the assigned task, they have to provide ratings based on (1) perceived task success and (2) post-dialogue self-reported user sentiment.
Perceived task success is measured as a binary answer on whether the system has provided all information, while the self-reported user sentiment is rated on a 5-point Likert scale ranging from very negative to very positive (see Appendix \ref{appendix:survey}).
After excluding low-quality samples using heuristic criteria, including dialogue length and linguistic diversity, we retained 86, 98, and 99 valid ratings for EmoLoop, LUSTER-base, and LUSTER, respectively.

We also note a clear difference in interaction style between human participants and the user simulator. 
In the paid human study, users tended to issue much shorter and more concise requests aimed at completing the assigned goal efficiently, whereas the simulator more often produced fuller, more explicit utterances (see Appendix~\ref{appendix:excerpts}). 
This distinction is important for interpreting the qualitative differences observed in system behaviour under human versus simulated evaluation.

\begin{table}[h]
\small
\centering
\begin{tabular}{lcc}
\toprule
 & Success ($\uparrow$) & Sentiment ($\uparrow$) \\
\midrule
EmoLoop & 0.826 & 3.86 \\
LUSTER-base & $\dagger$0.878 & 4.05 \\
LUSTER & $\dagger$\textbf{0.909} & $\dagger$\textbf{4.22} \\
\bottomrule
\end{tabular}
\caption{
    Human rating for task \textbf{success} and self-reported post-dialogue \textbf{sentiment}. Best scores across all systems are bolded. $\dagger$ denotes statistically significant improvement (p<0.05) from EmoLoop.
    }
    \label{tab:human-trial}
\end{table}

Table~\ref{tab:human-trial} shows that LUSTER achieves the highest ratings on both metrics.
Compared to LUSTER-base, incorporating emotion modelling and RL optimisation leads to consistent improvements in user-perceived task success and interaction quality.
LUSTER also outperforms EmoLoop significantly, suggesting that end-to-end emotional optimisation provides advantages over modular emotion integration and optimisation.

\subsubsection{Behavioural Analysis of Human Evaluation Logs}
\label{sec:behaviour-analysis}

The aggregate metrics show that affect-aware optimisation improves user sentiment, but they do not explain how system behaviour changes. 
To better understand this effect, we analyse behavioural patterns in the human-evaluation logs of LUSTER-base and LUSTER. 
We focus on five observable interaction behaviours: user frustration, user satisfaction, apology and repair, clarification, and warm closure. 
These categories are intended to capture surface-level interaction patterns rather than to provide an exhaustive annotation of emotional intelligence.

\begin{table}[h]
\centering
\footnotesize
\setlength\tabcolsep{5pt}
\begin{tabular}{lrr}
\toprule
Behaviour & LUSTER-base & LUSTER \\
\midrule
User satisfaction & 54.1\% & 69.7\% \\
Apology and repair & 18.9\% & 22.0\% \\
Clarification & 53.1\% & 72.7\% \\
Warm closure & 10.2\% & 22.2\% \\
\bottomrule
\end{tabular}
\caption{Behavioural indicators in the human evaluation logs. Statistics are first derived using rule-based filters and LLM-assisted inspection, and then verified by expert review.}
\label{tab:behaviour}
\end{table}

Table~\ref{tab:behaviour} suggests that LUSTER changes observable interaction behaviour in several ways. 
First, LUSTER produces substantially more clarification behaviour than LUSTER-base, suggesting that affect-aware optimisation encourages the system to ask for missing constraints or recover uncertain dialogue states rather than proceeding immediately. 
Second, apology and repair increase slightly. 
This indicates that the improvement is not simply a matter of apologising more frequently; rather, LUSTER more often combines affective acknowledgement with task-oriented recovery. 
Third, LUSTER leads to more warm closures with the user towards the end of the interaction, which is consistent with the higher proportion of satisfied users and the improvement in post-dialogue sentiment. 
Behavioural examples (over-apology, failure-triggered frustration, repair after confusion, and over-clarification) can be found in Appendix \ref{appendix:examples-behavioural-analysis}.

\subsubsection{LLM-based Preference Evaluation}

To further analyse nuanced behavioural differences that coarse automatic metrics or human ratings may not capture, we conducted a preference-based evaluation using LLMs between LUSTER and LUSTER-base.
Each system interacted with a user simulator under identical goals, and dialogue pairs were presented in random order to {GPT-4o-mini} (2025-04-16) and {DeepSeek-R1} (2025-05-28), which were tasked to select the better system (see Appendix~\ref{sec:appendix:evaluation-prompt}).
System identities were anonymised as System A and System B, and pair order was randomised, so judges did not know which output came from LUSTER.

\begin{figure}[ht]
    \centering
    \includegraphics[width=0.48\textwidth]{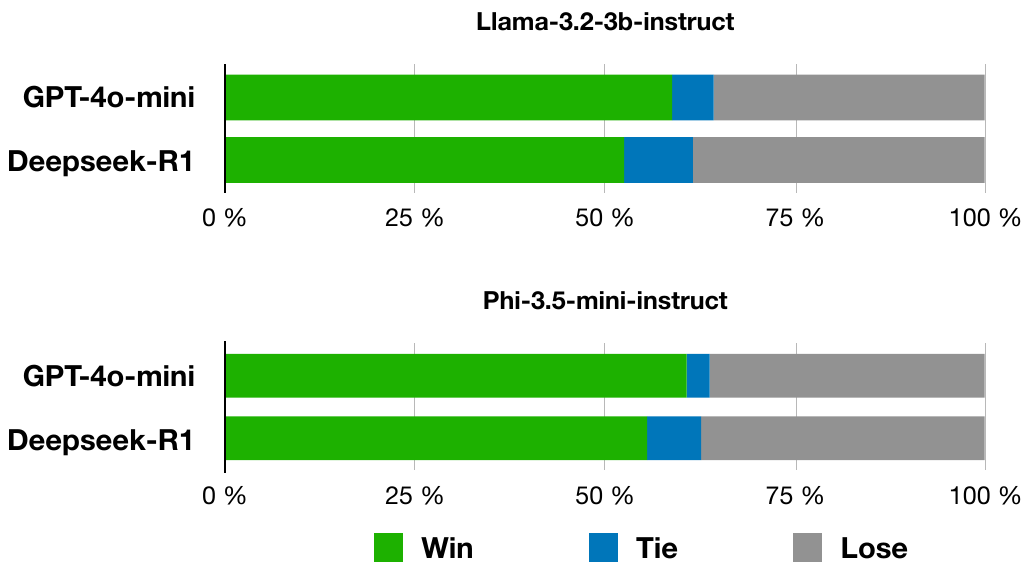}
    \caption{\small LLM preference of LUSTER against LUSTER-base using different backbones.}
    \label{fig:wr}
\end{figure}

Figure~\ref{fig:wr} presents the win-rate comparisons across 500 dialogues using Llama-3.2-3b-instruct and Phi-3.5-mini-instruct backbones.
Both LLM judges show a clear preference for LUSTER using both backbones: {GPT-4o-mini} assigns a strong majority of wins to LUSTER, while {DeepSeek-R1} also favours LUSTER with a more moderate margin.
These results provide additional evidence that emotional optimisation leads to nuanced yet perceptible improvements in conversational quality beyond task success alone.
Appendix~\ref{sec:appendix:case-study} illustrates representative cases in which LLM judges identify behavioural differences between systems.

\subsection{Scope, Limitations, and Future Directions}
LUSTER models coarse affective interaction signals rather than emotion in its full psychological complexity. The EmoWOZ user-emotion categories and EmoWOZ 2.0 system-conduct categories provide a structured way to incorporate affect into ToD optimisation, but they do not capture mixed emotions, appraisal, personality, long-term affective trajectories, or broader social context.
Similarly, our human evaluation measures perceived task success, overall performance, and self-reported post-dialogue sentiment; this score is affect-relevant, but should not be interpreted as a direct measure of emotion recognition or emotional intelligence.

These limitations define the scope of the work rather than undermine its contribution. 
Our goal is not to provide a complete psychological account of emotion, but to study whether structured affective signals can be integrated into end-to-end ToD optimisation and how they interact with task success, user experience, semantic expression, and observable dialogue behaviour. 
Our results show that affective optimisation is most useful when combined with task-oriented objectives, while also revealing where current affective modelling remains insufficient.

Our ablations isolate the effects of emotion modelling and reward design within a matched LUSTER architecture. 
They do not rule out the possibility that similar RL objectives could improve other ToD systems. 
However, applying the same RL procedure to large agentic systems or existing delexicalised end-to-end baselines is non-trivial: it would require different interaction interfaces, expanded lexical vocabularies, retrained embedding and output layers, and substantial additional computation. 
We therefore leave broader RL adaptation of other architectures to future work.

More broadly, LUSTER connects affect-aware ToD to richer user modelling. 
By estimating both the user's task-related state and affective state, the system models what the user is trying to achieve and how the user appears to feel during the interaction. 
These two dimensions are related to theory-of-mind-inspired accounts of interaction, but LUSTER does not yet claim to model the user's beliefs, intentions, or latent mental states in a psychologically complete sense. 
Future work could extend this connection by incorporating richer models of user belief, uncertainty, preference, and affective trajectory.

\section{Conclusion}
This work revisits the role of emotion in ToDs.
We show that affective signals can improve existing ToD systems, but their impact remains limited when they are integrated only locally, such as through module-level prediction or response-style control.

To address this, we introduced \textsc{LUSTER}, a lexicalised end-to-end ToD framework that integrates user-emotion modelling, system emotional-conduct prediction, and joint optimisation of task success and user sentiment. 
Experiments show that LUSTER achieves strong task performance, improved user sentiment, and low concept error under a noisy natural-language interaction protocol. 
Ablation studies further show that affective and task-oriented rewards play complementary roles, and backbone replication confirms that the main pattern is not tied to a single model family.

Human evaluation, behavioural analysis, and LLM-based preference judgements suggest that affect-aware optimisation leads to perceptible improvements in conversational quality, especially in clarification, repair, and affective closure. 
These findings show that emotional interaction is an important dimension of ToD performance: task success alone does not guarantee positive user experience, while affective style alone does not guarantee reliable task completion. 
\textsc{LUSTER} therefore provides a practical step toward ToD systems that jointly optimise task effectiveness and human-centric interaction quality.

In a broader context, this work signifies an important step towards incorporating an explicit theory of mind into the operation of LLMs. Theory of mind relates to modelling of cognitive and emotional states of others and, as such, represents an integral component of intelligence~\cite{Perner1999}. While LLMs do not by design model the theory of mind, they can imitate it quite well~\cite{chen-etal-2025-theory}. In this work, we show how modelling of user goal through dialogue states and modelling user affective state through emotion classes can be represented in an LLM backbone, both modelled by SL and RL and achieve overall better performance. This opens the door to the further development of such architectures in high-risk applications necessitating emotion modelling, such as psychotherapeutic contexts.

\section{Acknowledgements}
We thank the reviewers and the action editor for their insightful comments and suggested revisions, which significantly improved the quality of this paper. This work was funded by the European Research Council (ERC) under the Horizon 2020 research and innovation program (Grant No. STG2018 804636) and the Ministry of Culture and Science of North Rhine-Westphalia within the Lamarr Fellow Network. Computational resources were provided by the Centre for Information and Media Technology at Heinrich Heine University Düsseldorf and Google Cloud.

\bibliography{tacl2021}

@ARTICLE{pomdp,
  author={Young, Steve and Gašić, Milica and Thomson, Blaise and Williams, Jason D.},
  journal={Proceedings of the IEEE}, 
  title={POMDP-Based Statistical Spoken Dialog Systems: A Review}, 
  year={2013},
  volume={101},
  number={5},
  pages={1160-1179},
  doi={10.1109/JPROC.2012.2225812}}

@INPROCEEDINGS{online-rl-real-user,
  author={Gašić, M. and Breslin, C. and Henderson, M. and Kim, D. and Szummer, M. and Thomson, B. and Tsiakoulis, P. and Young, S.},
  booktitle={2013 IEEE International Conference on Acoustics, Speech and Signal Processing}, 
  title={On-line policy optimisation of Bayesian spoken dialogue systems via human interaction}, 
  year={2013},
  volume={},
  number={},
  pages={8367-8371},
  doi={10.1109/ICASSP.2013.6639297}}

@inproceedings{multiwoz,
    author = {Budzianowski, Pawe{\l} and Wen, Tsung-Hsien and Tseng, Bo-Hsiang  and Casanueva, I{\~n}igo and Ultes Stefan and Ramadan Osman and Ga{\v{s}}i\'c, Milica},
    title={MultiWOZ - A Large-Scale Multi-Domain Wizard-of-Oz Dataset for Task-Oriented Dialogue Modelling},
    booktitle={Proceedings of the 2018 Conference on Empirical Methods in Natural Language Processing (EMNLP)},
    year={2018}
}

@inproceedings{ertod,
    title = "From Chatter to Matter: Addressing Critical Steps of Emotion Recognition Learning in Task-oriented Dialogue",
    author = "Feng, Shutong  and
      Lubis, Nurul  and
      Ruppik, Benjamin  and
      Geishauser, Christian  and
      Heck, Michael  and
      Lin, Hsien-chin  and
      van Niekerk, Carel  and
      Vukovic, Renato  and
      Gasic, Milica",
    editor = "Stoyanchev, Svetlana  and
      Joty, Shafiq  and
      Schlangen, David  and
      Dusek, Ondrej  and
      Kennington, Casey  and
      Alikhani, Malihe",
    booktitle = "Proceedings of the 24th Annual Meeting of the Special Interest Group on Discourse and Dialogue",
    month = sep,
    year = "2023",
    address = "Prague, Czechia",
    publisher = "Association for Computational Linguistics",
    url = "https://aclanthology.org/2023.sigdial-1.8/",
    doi = "10.18653/v1/2023.sigdial-1.8",
    pages = "85--103"
}

@inproceedings{setsumbt,
    title = "Uncertainty Measures in Neural Belief Tracking and the Effects on Dialogue Policy Performance",
    author = "van Niekerk, Carel  and
      Malinin, Andrey  and
      Geishauser, Christian  and
      Heck, Michael  and
      Lin, Hsien-chin  and
      Lubis, Nurul  and
      Feng, Shutong  and
      Gasic, Milica",
    editor = "Moens, Marie-Francine  and
      Huang, Xuanjing  and
      Specia, Lucia  and
      Yih, Scott Wen-tau",
    booktitle = "Proceedings of the 2021 Conference on Empirical Methods in Natural Language Processing",
    month = nov,
    year = "2021",
    address = "Online and Punta Cana, Dominican Republic",
    publisher = "Association for Computational Linguistics",
    url = "https://aclanthology.org/2021.emnlp-main.623/",
    doi = "10.18653/v1/2021.emnlp-main.623",
    pages = "7901--7914"
}

@inproceedings{emoloop,
    title = "Infusing Emotions into Task-oriented Dialogue Systems: Understanding, Management, and Generation",
    author = "Feng, Shutong  and
      Lin, Hsien-chin  and
      Geishauser, Christian  and
      Lubis, Nurul  and
      van Niekerk, Carel  and
      Heck, Michael  and
      Ruppik, Benjamin Matthias  and
      Vukovic, Renato  and
      Gasic, Milica",
    editor = "Kawahara, Tatsuya  and
      Demberg, Vera  and
      Ultes, Stefan  and
      Inoue, Koji  and
      Mehri, Shikib  and
      Howcroft, David  and
      Komatani, Kazunori",
    booktitle = "Proceedings of the 25th Annual Meeting of the Special Interest Group on Discourse and Dialogue",
    month = sep,
    year = "2024",
    address = "Kyoto, Japan",
    publisher = "Association for Computational Linguistics",
    url = "https://aclanthology.org/2024.sigdial-1.60/",
    doi = "10.18653/v1/2024.sigdial-1.60",
    pages = "699--717"
}

@inproceedings{e2e-trainable-network,
    title = "A Network-based End-to-End Trainable Task-oriented Dialogue System",
    author = "Wen, Tsung-Hsien  and
      Vandyke, David  and
      Mrk{\v{s}}i{\'c}, Nikola  and
      Ga{\v{s}}i{\'c}, Milica  and
      Rojas-Barahona, Lina M.  and
      Su, Pei-Hao  and
      Ultes, Stefan  and
      Young, Steve",
    editor = "Lapata, Mirella  and
      Blunsom, Phil  and
      Koller, Alexander",
    booktitle = "Proceedings of the 15th Conference of the {E}uropean Chapter of the Association for Computational Linguistics: Volume 1, Long Papers",
    month = apr,
    year = "2017",
    address = "Valencia, Spain",
    publisher = "Association for Computational Linguistics",
    url = "https://aclanthology.org/E17-1042/",
    pages = "438--449"
}

@inproceedings{simpletod,
author = {Hosseini-Asl, Ehsan and McCann, Bryan and Wu, Chien-Sheng and Yavuz, Semih and Socher, Richard},
title = {A simple language model for task-oriented dialogue},
year = {2020},
isbn = {9781713829546},
publisher = {Curran Associates Inc.},
address = {Red Hook, NY, USA},
booktitle = {Proceedings of the 34th International Conference on Neural Information Processing Systems},
articleno = {1694},
numpages = {13},
location = {Vancouver, BC, Canada},
series = {NIPS '20}
}

@article{soloist,
    title = "Soloist: Building Task Bots at Scale with Transfer Learning and Machine Teaching",
    author = "Peng, Baolin  and
      Li, Chunyuan  and
      Li, Jinchao  and
      Shayandeh, Shahin  and
      Liden, Lars  and
      Gao, Jianfeng",
    editor = "Roark, Brian  and
      Nenkova, Ani",
    journal = "Transactions of the Association for Computational Linguistics",
    volume = "9",
    year = "2021",
    address = "Cambridge, MA",
    publisher = "MIT Press",
    url = "https://aclanthology.org/2021.tacl-1.49/",
    doi = "10.1162/tacl_a_00399",
    pages = "807--824"
}

@inproceedings{emowoz,
    title = "{E}mo{WOZ}: A Large-Scale Corpus and Labelling Scheme for Emotion Recognition in Task-Oriented Dialogue Systems",
    author = "Feng, Shutong  and
      Lubis, Nurul  and
      Geishauser, Christian  and
      Lin, Hsien-chin  and
      Heck, Michael  and
      van Niekerk, Carel  and
      Gasic, Milica",
    editor = "Calzolari, Nicoletta  and
      B{\'e}chet, Fr{\'e}d{\'e}ric  and
      Blache, Philippe  and
      Choukri, Khalid  and
      Cieri, Christopher  and
      Declerck, Thierry  and
      Goggi, Sara  and
      Isahara, Hitoshi  and
      Maegaard, Bente  and
      Mariani, Joseph  and
      Mazo, H{\'e}l{\`e}ne  and
      Odijk, Jan  and
      Piperidis, Stelios",
    booktitle = "Proceedings of the Thirteenth Language Resources and Evaluation Conference",
    month = jun,
    year = "2022",
    address = "Marseille, France",
    publisher = "European Language Resources Association",
    url = "https://aclanthology.org/2022.lrec-1.436/",
    pages = "4096--4113"
}

@inproceedings{convlab-3,
    title = "{C}onv{L}ab-3: A Flexible Dialogue System Toolkit Based on a Unified Data Format",
    author = "Zhu, Qi  and
      Geishauser, Christian  and
      Lin, Hsien-chin  and
      van Niekerk, Carel  and
      Peng, Baolin  and
      Zhang, Zheng  and
      Feng, Shutong  and
      Heck, Michael  and
      Lubis, Nurul  and
      Wan, Dazhen  and
      Zhu, Xiaochen  and
      Gao, Jianfeng  and
      Gasic, Milica  and
      Huang, Minlie",
    editor = "Feng, Yansong  and
      Lefever, Els",
    booktitle = "Proceedings of the 2023 Conference on Empirical Methods in Natural Language Processing: System Demonstrations",
    month = dec,
    year = "2023",
    address = "Singapore",
    publisher = "Association for Computational Linguistics",
    url = "https://aclanthology.org/2023.emnlp-demo.9/",
    doi = "10.18653/v1/2023.emnlp-demo.9",
    pages = "106--123"
}

@techreport{phi-3.5,
author = {Abdin, Marah I and Ade Jacobs, Sam and Awan, Ammar Ahmad and Aneja, Jyoti and Awadallah, Ahmed and others},
title = {Phi-3 Technical Report: A Highly Capable Language Model Locally on Your Phone},
institution = {Microsoft},
year = {2024},
month = {August},
url = {https://www.microsoft.com/en-us/research/publication/phi-3-technical-report-a-highly-capable-language-model-locally-on-your-phone/},
number = {MSR-TR-2024-12},
}

@inproceedings{lava,
    title = "{LAVA}: Latent Action Spaces via Variational Auto-encoding for Dialogue Policy Optimization",
    author = "Lubis, Nurul  and
      Geishauser, Christian  and
      Heck, Michael  and
      Lin, Hsien-chin  and
      Moresi, Marco  and
      van Niekerk, Carel  and
      Gasic, Milica",
    editor = "Scott, Donia  and
      Bel, Nuria  and
      Zong, Chengqing",
    booktitle = "Proceedings of the 28th International Conference on Computational Linguistics",
    month = dec,
    year = "2020",
    address = "Barcelona, Spain (Online)",
    publisher = "International Committee on Computational Linguistics",
    url = "https://aclanthology.org/2020.coling-main.41/",
    doi = "10.18653/v1/2020.coling-main.41",
    pages = "465--479"
}

@inproceedings{gpt-critic,
title={{GPT}-Critic: Offline Reinforcement Learning for End-to-End Task-Oriented Dialogue Systems},
author={Youngsoo Jang and Jongmin Lee and Kee-Eung Kim},
booktitle={International Conference on Learning Representations},
year={2022},
url={https://openreview.net/forum?id=qaxhBG1UUaS}
}

@inproceedings{simplellama,
  TITLE = {{A Unified Approach to Emotion Detection and Task-Oriented Dialogue Modeling}},
  AUTHOR = {Stricker, Armand and Paroubek, Patrick},
  URL = {https://hal.science/hal-04415809},
  BOOKTITLE = {{IWSDS}},
  ADDRESS = {Sapporo (Japon), Japan},
  YEAR = {2024},
  MONTH = Mar,
  HAL_ID = {hal-04415809},
  HAL_VERSION = {v1},
}

@inproceedings{chatgpt-dst,
    title = "{C}hat{GPT} for Zero-shot Dialogue State Tracking: A Solution or an Opportunity?",
    author = "Heck, Michael  and
      Lubis, Nurul  and
      Ruppik, Benjamin  and
      Vukovic, Renato  and
      Feng, Shutong  and
      Geishauser, Christian  and
      Lin, Hsien-chin  and
      van Niekerk, Carel  and
      Gasic, Milica",
    editor = "Rogers, Anna  and
      Boyd-Graber, Jordan  and
      Okazaki, Naoaki",
    booktitle = "Proceedings of the 61st Annual Meeting of the Association for Computational Linguistics (Volume 2: Short Papers)",
    month = jul,
    year = "2023",
    address = "Toronto, Canada",
    publisher = "Association for Computational Linguistics",
    url = "https://aclanthology.org/2023.acl-short.81/",
    doi = "10.18653/v1/2023.acl-short.81",
    pages = "936--950"
}

@inproceedings{chatgpt-erc,
    title = "Affect Recognition in Conversations Using Large Language Models",
    author = "Feng, Shutong  and
      Sun, Guangzhi  and
      Lubis, Nurul  and
      Wu, Wen  and
      Zhang, Chao  and
      Gasic, Milica",
    editor = "Kawahara, Tatsuya  and
      Demberg, Vera  and
      Ultes, Stefan  and
      Inoue, Koji  and
      Mehri, Shikib  and
      Howcroft, David  and
      Komatani, Kazunori",
    booktitle = "Proceedings of the 25th Annual Meeting of the Special Interest Group on Discourse and Dialogue",
    month = sep,
    year = "2024",
    address = "Kyoto, Japan",
    publisher = "Association for Computational Linguistics",
    url = "https://aclanthology.org/2024.sigdial-1.23/",
    doi = "10.18653/v1/2024.sigdial-1.23",
    pages = "259--273"
}

@inproceedings{ontology-construction-cot,
    title = "Dialogue Ontology Relation Extraction via Constrained Chain-of-Thought Decoding",
    author = "Vukovic, Renato  and
      Arps, David  and
      van Niekerk, Carel  and
      Ruppik, Benjamin Matthias  and
      Lin, Hsien-chin  and
      Heck, Michael  and
      Gasic, Milica",
    editor = "Kawahara, Tatsuya  and
      Demberg, Vera  and
      Ultes, Stefan  and
      Inoue, Koji  and
      Mehri, Shikib  and
      Howcroft, David  and
      Komatani, Kazunori",
    booktitle = "Proceedings of the 25th Annual Meeting of the Special Interest Group on Discourse and Dialogue",
    month = sep,
    year = "2024",
    address = "Kyoto, Japan",
    publisher = "Association for Computational Linguistics",
    url = "https://aclanthology.org/2024.sigdial-1.33/",
    doi = "10.18653/v1/2024.sigdial-1.33",
    pages = "370--384"
}

@inproceedings{prefix-tuning,
    title = "Prefix-Tuning: Optimizing Continuous Prompts for Generation",
    author = "Li, Xiang Lisa  and
      Liang, Percy",
    editor = "Zong, Chengqing  and
      Xia, Fei  and
      Li, Wenjie  and
      Navigli, Roberto",
    booktitle = "Proceedings of the 59th Annual Meeting of the Association for Computational Linguistics and the 11th International Joint Conference on Natural Language Processing (Volume 1: Long Papers)",
    month = aug,
    year = "2021",
    address = "Online",
    publisher = "Association for Computational Linguistics",
    url = "https://aclanthology.org/2021.acl-long.353/",
    doi = "10.18653/v1/2021.acl-long.353",
    pages = "4582--4597"
}

@inproceedings{lora,
title={Lo{RA}: Low-Rank Adaptation of Large Language Models},
author={Edward J Hu and yelong shen and Phillip Wallis and Zeyuan Allen-Zhu and Yuanzhi Li and Shean Wang and Lu Wang and Weizhu Chen},
booktitle={International Conference on Learning Representations},
year={2022},
url={https://openreview.net/forum?id=nZeVKeeFYf9}
}

@Book{speech-language-processing,
  author =       "Daniel Jurafsky and James H. Martin",
  title =        "Speech and Language Processing: An Introduction to
                 Natural Language Processing, Computational Linguistics,
                 and Speech Recognition with Language Models",
  year =         "2025",
  url = {https://web.stanford.edu/~jurafsky/slp3/},
  note = "Online manuscript released January 12, 2025",
  edition =         "3rd",
  }

@inproceedings{instructgpt,
author = {Ouyang, Long and Wu, Jeff and Jiang, Xu and Almeida, Diogo and others},
title = {Training language models to follow instructions with human feedback},
year = {2022},
isbn = {9781713871088},
publisher = {Curran Associates Inc.},
address = {Red Hook, NY, USA},
booktitle = {Proceedings of the 36th International Conference on Neural Information Processing Systems},
articleno = {2011},
numpages = {15},
location = {New Orleans, LA, USA},
series = {NIPS '22}
}

@inproceedings{chatgpt-sl,
    title = "Speech-based Slot Filling using Large Language Models",
    author = "Sun, Guangzhi  and
      Feng, Shutong  and
      Jiang, Dongcheng  and
      Zhang, Chao  and
      Gasic, Milica  and
      Woodland, Phil",
    editor = "Ku, Lun-Wei  and
      Martins, Andre  and
      Srikumar, Vivek",
    booktitle = "Findings of the Association for Computational Linguistics: ACL 2024",
    month = aug,
    year = "2024",
    address = "Bangkok, Thailand",
    publisher = "Association for Computational Linguistics",
    url = "https://aclanthology.org/2024.findings-acl.379/",
    doi = "10.18653/v1/2024.findings-acl.379",
    pages = "6351--6362"
}

@inproceedings{emous,
author = {Lin, Hsien-Chin and Feng, Shutong and Geishauser, Christian and Lubis, Nurul and van Niekerk, Carel and Heck, Michael and Ruppik, Benjamin and Vukovic, Renato and Gasi\'{c}, Milica},
title = {EmoUS: Simulating User Emotions in Task-Oriented Dialogues},
year = {2023},
isbn = {9781450394086},
publisher = {Association for Computing Machinery},
address = {New York, NY, USA},
url = {https://doi.org/10.1145/3539618.3592092},
doi = {10.1145/3539618.3592092},
booktitle = {Proceedings of the 46th International ACM SIGIR Conference on Research and Development in Information Retrieval},
pages = {2526–2531},
numpages = {6},
location = {Taipei, Taiwan},
series = {SIGIR '23}
}

@inproceedings{pg,
author = {Sutton, Richard S. and McAllester, David and Singh, Satinder and Mansour, Yishay},
title = {Policy gradient methods for reinforcement learning with function approximation},
year = {1999},
publisher = {MIT Press},
address = {Cambridge, MA, USA},
booktitle = {Proceedings of the 13th International Conference on Neural Information Processing Systems},
pages = {1057–1063},
numpages = {7},
location = {Denver, CO},
series = {NIPS'99}
}

@misc{ppo,
      title={Proximal Policy Optimization Algorithms}, 
      author={John Schulman and Filip Wolski and Prafulla Dhariwal and Alec Radford and Oleg Klimov},
      year={2017},
      eprint={1707.06347},
      archivePrefix={arXiv},
      primaryClass={cs.LG},
      url={https://arxiv.org/abs/1707.06347}, 
}

@inproceedings{archer,
title={Ar{CH}er: Training Language Model Agents via Hierarchical Multi-Turn {RL}},
author={Yifei Zhou and Andrea Zanette and Jiayi Pan and Sergey Levine and Aviral Kumar},
booktitle={Forty-first International Conference on Machine Learning},
year={2024},
url={https://openreview.net/forum?id=b6rA0kAHT1}
}

@inproceedings{albert,
title={ALBERT: A Lite BERT for Self-supervised Learning of Language Representations},
author={Zhenzhong Lan and Mingda Chen and Sebastian Goodman and Kevin Gimpel and Piyush Sharma and Radu Soricut},
booktitle={International Conference on Learning Representations},
year={2020},
url={https://openreview.net/forum?id=H1eA7AEtvS}
}

@inproceedings{ddpt,
    title = "Dynamic Dialogue Policy for Continual Reinforcement Learning",
    author = "Geishauser, Christian  and
      van Niekerk, Carel  and
      Lin, Hsien-chin  and
      Lubis, Nurul  and
      Heck, Michael  and
      Feng, Shutong  and
      Ga{\v{s}}i{\'c}, Milica",
    editor = "Calzolari, Nicoletta  and
      Huang, Chu-Ren  and
      Kim, Hansaem  and
      Pustejovsky, James  and
      Wanner, Leo  and
      Choi, Key-Sun  and
      Ryu, Pum-Mo  and
      Chen, Hsin-Hsi  and
      Donatelli, Lucia  and
      Ji, Heng  and
      Kurohashi, Sadao  and
      Paggio, Patrizia  and
      Xue, Nianwen  and
      Kim, Seokhwan  and
      Hahm, Younggyun  and
      He, Zhong  and
      Lee, Tony Kyungil  and
      Santus, Enrico  and
      Bond, Francis  and
      Na, Seung-Hoon",
    booktitle = "Proceedings of the 29th International Conference on Computational Linguistics",
    month = oct,
    year = "2022",
    address = "Gyeongju, Republic of Korea",
    publisher = "International Committee on Computational Linguistics",
    url = "https://aclanthology.org/2022.coling-1.21/",
    pages = "266--284"
}

@misc{gpt4,
      title={GPT-4 Technical Report}, 
      author={OpenAI},
      year={2023},
      eprint={2303.08774},
      archivePrefix={arXiv},
      primaryClass={cs.CL}
}

@misc{llama2,
      title={Llama 2: Open Foundation and Fine-Tuned Chat Models}, 
      author={Hugo Touvron and Louis Martin and Kevin Stone and Peter Albert and Amjad Almahairi and others},
      year={2023},
      eprint={2307.09288},
      archivePrefix={arXiv},
      primaryClass={cs.CL}
}

@misc{deepseek,
      title={DeepSeek-R1: Incentivizing Reasoning Capability in LLMs via Reinforcement Learning}, 
      author={DeepSeek-AI},
      year={2025},
      eprint={2501.12948},
      archivePrefix={arXiv},
      primaryClass={cs.CL},
      url={https://arxiv.org/abs/2501.12948}, 
}

@book{affective-computing,
  address = {Cambridge, MA},
  author = {Picard, Rosalind W.},
  description = {2000 Paperback 978-0-262-66115-7},
  isbn = {978-0-262-16170-1},
  publisher = {MIT Press},
  title = {Affective Computing},
  username = {flint63},
  year = 1997
}

@inproceedings{react,
  title = {{ReAct}: Synergizing Reasoning and Acting in Language Models},
  author = {Yao, Shunyu and Zhao, Jeffrey and Yu, Dian and Du, Nan and Shafran, Izhak and Narasimhan, Karthik and Cao, Yuan},
  booktitle = {International Conference on Learning Representations (ICLR) },
  year = {2023},
  html = {https://arxiv.org/abs/2210.03629},
}

@inproceedings{cot,
author = {Wei, Jason and Wang, Xuezhi and Schuurmans, Dale and Bosma, Maarten and Ichter, Brian and Xia, Fei and Chi, Ed H. and Le, Quoc V. and Zhou, Denny},
title = {Chain-of-thought prompting elicits reasoning in large language models},
year = {2022},
isbn = {9781713871088},
publisher = {Curran Associates Inc.},
address = {Red Hook, NY, USA},
booktitle = {Proceedings of the 36th International Conference on Neural Information Processing Systems},
articleno = {1800},
numpages = {14},
location = {New Orleans, LA, USA},
series = {NIPS '22}
}

@inproceedings{polyak,
  author = {Haarnoja, Tuomas and Zhou, Aurick and Abbeel, Pieter and Levine, Sergey},
  booktitle = {ICML},
  editor = {Dy, Jennifer G. and Krause, Andreas},
  ee = {http://proceedings.mlr.press/v80/haarnoja18b.html},
  pages = {1856-1865},
  publisher = {PMLR},
  series = {Proceedings of Machine Learning Research},
  title = {Soft Actor-Critic: Off-Policy Maximum Entropy Deep Reinforcement Learning with a Stochastic Actor.},
  url = {http://dblp.uni-trier.de/db/conf/icml/icml2018.html#HaarnojaZAL18},
  volume = 80,
  year = 2018
}

@inproceedings{reservoir-sampling,
author = {Isele, David and Cosgun, Akansel},
title = {Selective experience replay for lifelong learning},
year = {2018},
isbn = {978-1-57735-800-8},
publisher = {AAAI Press},
booktitle = {Proceedings of the Thirty-Second AAAI Conference on Artificial Intelligence and Thirtieth Innovative Applications of Artificial Intelligence Conference and Eighth AAAI Symposium on Educational Advances in Artificial Intelligence},
articleno = {404},
numpages = {8},
location = {New Orleans, Louisiana, USA},
series = {AAAI'18/IAAI'18/EAAI'18}
}

@inproceedings{importance-sampling,
author = {Precup, Doina and Sutton, Richard S. and Singh, Satinder P.},
title = {Eligibility Traces for Off-Policy Policy Evaluation},
year = {2000},
isbn = {1558607072},
publisher = {Morgan Kaufmann Publishers Inc.},
address = {San Francisco, CA, USA},
booktitle = {Proceedings of the Seventeenth International Conference on Machine Learning},
pages = {759–766},
numpages = {8},
series = {ICML '00}
}

@inproceedings{fnctod,
    title = "Large Language Models as Zero-shot Dialogue State Tracker through Function Calling",
    author = "Li, Zekun  and
      Chen, Zhiyu  and
      Ross, Mike  and
      Huber, Patrick  and
      Moon, Seungwhan  and
      Lin, Zhaojiang  and
      Dong, Xin  and
      Sagar, Adithya  and
      Yan, Xifeng  and
      Crook, Paul",
    editor = "Ku, Lun-Wei  and
      Martins, Andre  and
      Srikumar, Vivek",
    booktitle = "Proceedings of the 62nd Annual Meeting of the Association for Computational Linguistics (Volume 1: Long Papers)",
    month = aug,
    year = "2024",
    address = "Bangkok, Thailand",
    publisher = "Association for Computational Linguistics",
    url = "https://aclanthology.org/2024.acl-long.471/",
    doi = "10.18653/v1/2024.acl-long.471",
    pages = "8688--8704"
}

@inproceedings{llms-all-you-need-tod,
    title = "Are Large Language Models All You Need for Task-Oriented Dialogue?",
    author = "Hude{\v{c}}ek, Vojt{\v{e}}ch  and
      Dusek, Ondrej",
    editor = "Stoyanchev, Svetlana  and
      Joty, Shafiq  and
      Schlangen, David  and
      Dusek, Ondrej  and
      Kennington, Casey  and
      Alikhani, Malihe",
    booktitle = "Proceedings of the 24th Annual Meeting of the Special Interest Group on Discourse and Dialogue",
    month = sep,
    year = "2023",
    address = "Prague, Czechia",
    publisher = "Association for Computational Linguistics",
    url = "https://aclanthology.org/2023.sigdial-1.21/",
    doi = "10.18653/v1/2023.sigdial-1.21",
    pages = "216--228"
}

@article{gpt2,
  title={Language Models are Unsupervised Multitask Learners},
  author={Radford, Alec and Wu, Jeff and Child, Rewon and Luan, David and Amodei, Dario and Sutskever, Ilya},
  year={2019}
}

@inproceedings{rlhf,
author = {Ouyang, Long and Wu, Jeff and Jiang, Xu and Almeida, Diogo and Wainwright, Carroll L. and others},
title = {Training language models to follow instructions with human feedback},
year = {2022},
isbn = {9781713871088},
publisher = {Curran Associates Inc.},
address = {Red Hook, NY, USA},
booktitle = {Proceedings of the 36th International Conference on Neural Information Processing Systems},
articleno = {2011},
numpages = {15},
location = {New Orleans, LA, USA},
series = {NIPS '22}
}

@misc{qwen35,
    title = {Qwen3.5: Accelerating Productivity with Native Multimodal Agents},
    url = {https://qwen.ai/blog?id=qwen3.5},
    author = {Qwen Team},
    month = {February},
    year = {2026}
}

@inproceedings{young2002talking,
  title={Talking to machines (statistically speaking).},
  author={Young, Steve J},
  booktitle={INTERSPEECH},
  pages={9--16},
  year={2002}
}

@article{emotion-as-state,
  title={An emotion-sensitive dialogue policy for task-oriented dialogue system},
  author={Zhu, Hui and Wang, Xv and Wang, Zhenyu and Xv, Kai},
  journal={Scientific Reports},
  volume={14},
  number={1},
  pages={19759},
  year={2024},
  publisher={Nature Publishing Group UK London}
}

@article{emotion-as-reward,
title = {Emotion-sensitive deep dyna-Q learning for task-completion dialogue policy learning},
journal = {Neurocomputing},
volume = {459},
pages = {122-130},
year = {2021},
issn = {0925-2312},
doi = {https://doi.org/10.1016/j.neucom.2021.06.075},
url = {https://www.sciencedirect.com/science/article/pii/S0925231221010055},
author = {Rui Zhang and Zhenyu Wang and Mengdan Zheng and Yangyang Zhao and Zhenhua Huang},
}

@misc{llama3,
      title={The Llama 3 Herd of Models}, 
      author={Aaron Grattafiori and Abhimanyu Dubey and Abhinav Jauhri and Abhinav Pandey and others},
      year={2024},
      eprint={2407.21783},
      archivePrefix={arXiv},
      primaryClass={cs.AI},
      url={https://arxiv.org/abs/2407.21783}, 
}

@inproceedings{dialcrowd,
    title = "{D}ial{C}rowd 2.0: A Quality-Focused Dialog System Crowdsourcing Toolkit",
    author = "Huynh, Jessica  and
      Chiang, Ting-Rui  and
      Bigham, Jeffrey  and
      Eskenazi, Maxine",
    editor = "Calzolari, Nicoletta  and
      B{\'e}chet, Fr{\'e}d{\'e}ric  and
      Blache, Philippe  and
      Choukri, Khalid  and
      Cieri, Christopher  and
      Declerck, Thierry  and
      Goggi, Sara  and
      Isahara, Hitoshi  and
      Maegaard, Bente  and
      Mariani, Joseph  and
      Mazo, H{\'e}l{\`e}ne  and
      Odijk, Jan  and
      Piperidis, Stelios",
    booktitle = "Proceedings of the Thirteenth Language Resources and Evaluation Conference",
    month = jun,
    year = "2022",
    address = "Marseille, France",
    publisher = "European Language Resources Association",
    url = "https://aclanthology.org/2022.lrec-1.134/",
    pages = "1256--1263",
}

@inproceedings{gentus,
    title = "{G}en{TUS}: Simulating User Behaviour and Language in Task-oriented Dialogues with Generative Transformers",
    author = "Lin, Hsien-chin  and
      Geishauser, Christian  and
      Feng, Shutong  and
      Lubis, Nurul  and
      van Niekerk, Carel  and
      Heck, Michael  and
      Gasic, Milica",
    editor = "Lemon, Oliver  and
      Hakkani-Tur, Dilek  and
      Li, Junyi Jessy  and
      Ashrafzadeh, Arash  and
      Garcia, Daniel Hern{\'a}ndez  and
      Alikhani, Malihe  and
      Vandyke, David  and
      Du{\v{s}}ek, Ond{\v{r}}ej",
    booktitle = "Proceedings of the 23rd Annual Meeting of the Special Interest Group on Discourse and Dialogue",
    month = sep,
    year = "2022",
    address = "Edinburgh, UK",
    publisher = "Association for Computational Linguistics",
    url = "https://aclanthology.org/2022.sigdial-1.28/",
    doi = "10.18653/v1/2022.sigdial-1.28",
    pages = "270--282",
}

@incollection{Perner1999,
  author    = {Josef Perner},
  title     = {Theory of Mind},
  booktitle = {Developmental Psychology: Achievements and Prospects},
  editor     = {Michael Bennett},
  pages      = {205--230},
  year       = {1999},
  publisher  = {Psychology Press}
}

@inproceedings{chen-etal-2025-theory,
    title = "Theory of Mind in Large Language Models: Assessment and Enhancement",
    author = "Chen, Ruirui  and
      Jiang, Weifeng  and
      Qin, Chengwei  and
      Tan, Cheston",
    editor = "Che, Wanxiang  and
      Nabende, Joyce  and
      Shutova, Ekaterina  and
      Pilehvar, Mohammad Taher",
    booktitle = "Proceedings of the 63rd Annual Meeting of the Association for Computational Linguistics (Volume 1: Long Papers)",
    month = jul,
    year = "2025",
    address = "Vienna, Austria",
    publisher = "Association for Computational Linguistics",
    url = "https://aclanthology.org/2025.acl-long.1522/",
    doi = "10.18653/v1/2025.acl-long.1522",
    pages = "31539--31558",
    ISBN = "979-8-89176-251-0"
}
\bibliographystyle{acl_natbib}

\onecolumn

\appendix
\counterwithin{equation}{section}
\counterwithin{figure}{section}
\counterwithin{table}{section}
\section{Supplementary Information}
\subsection{Computing Resources}
\label{appendix:sec:compute}
All modular systems were trained and deployed using a single NVIDIA RTX 2080 (8GB) graphics processing unit (GPU). The end-to-end systems {SimpleLLAMA} and {EmoLLAMA} were trained and evaluated on an NVIDIA A100 (80GB) GPU. {LUSTER} variants were trained on a single NVIDIA A100 (40GB) GPU, with inference requiring at least an NVIDIA RTX 6000 (24GB) GPU. While training {LUSTER} demands more computational resources than modular systems, it remains more resource-efficient than LLAMA-based end-to-end models.

\subsection{Proximal Policy Optimisation for Single-turn RL}
\label{sec:appendix:ppo}

The advantage function $A_\pi(s,a)$ denotes the difference between the action-value function $Q_\pi(s,a)$, which is the expected return when taking action \(a\) in state \(s\) and from then on following the policy $\pi$ and the state value function $V_\pi(s)$, which is the expected return of state \(s\) for policy $\pi$. 
\begin{align}
  A_{\pi}(s,a) &=Q_{\pi}(s,a)-V_\pi(s)
\end{align}
In the context of single-turn response generation with an episode length of $L$ (Equation \ref{eqn:turn-level-state-1}) for the $t$-th interaction of a dialogue with a policy parameterised with $\theta$,
\begin{align}
  Q_{\pi_\theta}(s,a) &=\mathbb{E}_{\pi_\theta}\sum^{L-l-1}_{k=0}[\gamma^kr^{k+1}_{t}|s=s_t^{l},a=a_t^{l}] \\
  V_{\pi_\theta}(s) &=\mathbb{E}_{\pi_\theta}\sum^{L-l-1}_{k=0}[\gamma^{k}r_t^{k+1}|s=s_t^l] \\
  \hat{A}_t^l &= Q_{\pi_\theta}(s,a) - V_{\pi_\theta}(s)
\end{align}

The surrogate loss consisting of the clipping term and the KL-penalty term is given by:
\begin{align}
    \begin{split}
        \mathcal{L}_{\text{PPO}} = \mathbb{E}_t \Big[
        &\min \Big(
        \frac{\pi_\theta(a_t^l \mid s_t^l)}{\pi_{\theta_{\text{old}}}(a_t^l \mid s_t^l)} \hat{A}_t^l, \ \text{clip}\left( \frac{\pi_\theta(a_t^l \mid s_t^l)}{\pi_{\theta_{\text{old}}}(a_t^l \mid s_t^l)}, 1 - \epsilon,\ 1 + \epsilon \right) \hat{A}_t^l
        \Big) \\
        & - \beta \cdot \sum_{a} \pi_{\theta_{\text{old}}}(a \mid s_t^l) \log \left( \frac{\pi_{\theta_{\text{old}}}(a \mid s_t^l)}{\pi_\theta(a \mid s_t^l)} \right)
        \Big],
    \end{split}
    \label{eqn:surrogate_loss}
\end{align}
where $\text{clip}(x,a,b)$ ensures that $x$ is within the interval $[a,b]$.
The gradient of the surrogate loss (Eq~\ref{eqn:surrogate_loss}) approximates the direction of the policy gradient~(Eqn~\ref{eqn:pg}). 

To estimate advantage function $\hat{A}_t$ needed for the surrogate loss, generalised advantage estimation is used:
\begin{align}
    \begin{split}
    \hat{A}_t^l = \sum_{j=0}^{L - j - 1} (\gamma_{\text{turn}} \lambda)^j \, \delta_t^{l + j}
    \quad \text{where}
    \quad
    \delta_t^l &= r_t^l + \gamma_{\text{turn}} V_\pi(s_{t+1}^l) - V_{\pi_\theta}(s_t^l)
    \end{split} \label{eqn:ppo-adv}
\end{align}
Parameter $\lambda \in [0,1]$ trades off bias (lower $\lambda$) and variance (higher $\lambda$). $V_{\pi_\theta}$ is initialised with the reward model $R_\phi$ for complete sequences, and parameters $\phi$ are updated in such a way so that $V_{\pi_\theta}$ estimates observed returns for partial sequences. In this way, the critic parameterised with  $\phi$ influences the actor parameterised with $\theta$. We follow default hyperparameters in the PPO trainer of TRL library, which follows \citet{rlhf}.

PPO is an online learning algorithm, which means it needs to adopt exploration during the process of optimisation. That is achieved by applying a temperature parameter to the softmax layer of policy $\pi_\theta$ to induce more varied responses.

In practice, the critic is modelled by a single-layer value head attached to the output of the language model. It projects the logits of the token into a scalar. The input for calculating $V_{\pi_\theta}(s)$ is the logit of the last token in $s$, and that for calculating $Q_{\pi_\theta}(s,a)$ is the logit of $a$, since the action $a$ is the next token generated.

\newpage
\subsection{Dialogue-level Critics}
\label{sec:appendix:critic}

\subsubsection{Architecture}
\begin{figure*}[ht]
    \centering
    \begin{subfigure}{0.4\textwidth}
        \centering
        \includegraphics[width=\linewidth]{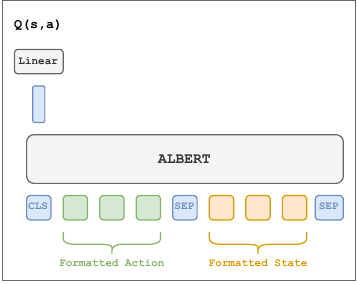}
        \caption{Action-state value function}
        \label{fig:critics-q}
    \end{subfigure}
    \begin{subfigure}{0.4\textwidth}
        \centering
        \includegraphics[width=\linewidth]{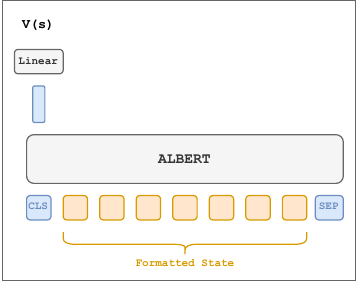}
        \caption{State value function}
        \label{fig:critics-v}
    \end{subfigure}
    \caption{Dialogue-level Critics}
    \label{fig:critics}
\end{figure*}

Figure \ref{fig:critics} illustrates the modelling of dialogue critics. We format the dialogue-level state $\mathcal{S}_t$ and action $\mathcal{A}_t$ from the generated sequence $s_t^0$ and $a_t^{1:L-1}$ to obtain a compact input to the critic. 

\paragraph{Textual State Representation} The input to the state value function is in the form of the following:\\

\textit{<domain> <slot> <``inactive''|``informed''|``unknown''> ; ...} [SEP]\textit{``N <domain> found''|``no entity found''|``database not available''} [SEP] \textit{``system :''} $u_{\sys,t-1}$ [SEP] \textit{``user :''} $u_{\usr,t}$\\

The first segment contains the system's belief about the user's goal. Instead of actual informed value, we limit the status of each slot to be one of the three: \textit{informed}, \textit{unknown} if not informed, or \textit{inactive} if the <domain> is not the active domain determined by the system $\text{<domain>}\neq d_t$.

The second segment contains a summary of the database query status. The status informs the number of entities returned $N$ for $N>0$ or informs ``no entity found'' if $N=0$. There is a special case when the active domain is ``general'' for small-talk and greetings. Under such circumstances, the status informs ``database not available''.

To account briefly for the dialogue history, we also include the previous system and user turns $u_{\sys,t-1}$ and $u_{\usr,t}$.

\paragraph{Textual Action Representation} The formatted action for the state-action value function is in the form of the following. Values are excluded.\\

\textit{<domain> <intent> <slot> ; ...}

\subsubsection{Training}

Following the objective functions defined in Equation \ref{eqn:obj-q} and \ref{eqn:obj-v}, for each critic update step:
\begin{align}
    \phi &\leftarrow \phi-\eta_Q\nabla J_\phi(Q) \\
    \psi &\leftarrow \psi-\eta_V\nabla J_\psi(V)
\end{align}

Parameters of target critics are then updated via Polyak averaging~\citep{polyak}:
\begin{align}
    \bar{\phi} &\leftarrow (1-\tau)\bar{\phi} + \tau \phi \label{eqn:polyak-q}\\
    \bar{\psi} &\leftarrow (1-\tau)\bar{\psi} + \tau \psi \label{eqn:polyak-v}
\end{align}

\newpage
\subsection{Pseudo Code}
\definecolor{darkgreen}{rgb}{0, 0.5, 0}
\begin{algorithm}[h]
\footnotesize
\begin{algorithmic}[1]
\State {Initialise parameters $\theta, \phi, \psi, \bar{\phi}, \bar{\psi}$}
\State \textcolor{darkgreen}{\#\# Actor Pretraining using EmoWOZ}
\For{each actor step}
\State $\theta \leftarrow \theta - \nabla \mathcal{L}_{\text{SL}}$\Comment{Equation \ref{eqn:loss-sl}}
\EndFor
\State \textcolor{darkgreen}{\#\# Critic Pretraining using EmoWOZ, not applicable when using emotion advantage only.}
\For{each critic step}
\State \textcolor{darkgreen}{\#\# Update utterance-level Q and V functions by target function bootstrapping.}
\State $\phi \leftarrow \phi - \eta_Q\nabla \mathcal{L}_Q$ \Comment{Equation~\ref{eqn:obj-q}}
\State $\psi \leftarrow \psi - \eta_V\nabla \mathcal{L}_V$ 
\Comment{Equation~\ref{eqn:obj-v}}
\State \textcolor{darkgreen}{\#\# Update target Q and V functions via Polyak averaging.}
\State $\bar{\phi} \leftarrow (1 - \tau)\bar{\phi} + \tau\phi$ \Comment{Equation~\ref{eqn:polyak-q}}
\State $\bar{\psi} \leftarrow (1 - \tau)\bar{\psi} + \tau\psi$ \Comment{Equation~\ref{eqn:polyak-v}}
\EndFor

\State \textcolor{darkgreen}{\#\# Initialise and fill experience replay buffer $\mathcal{D}$}
\For{each dialogue while buffer\_size not reached}
\For{each turn}
\State Execute $a_t^{0:L-1} \sim \pi_\theta(\cdot|s_t^0)$, obtain the next state $s_{t+1}^0$ and reward \{$r_t^L$, $A_t$, or $\mathcal{R}_{\text{combine},t}$\}, add to buffer $\mathcal{D}$.
\EndFor
\EndFor

\For{each iteration using buffer $\mathcal{D}$}
\State Repeat line 17-21 until $\mathcal{D}$ is full.
\State Repeat line 7-14 for critic update.
\For{each off-policy experience}
    \State Replay experience and sale reward with importance sampling weight.\Comment{Equation \ref{eqn:is-weight}}
\EndFor
\State \textcolor{darkgreen}{\#\# Update token-level actor with utterance-level critic.}
\For{each actor step}
\State $\theta \leftarrow \theta - \nabla \mathcal{L}_{\text{PPO}}$ \Comment{Equation~\ref{eqn:pg}, approximated with \ref{eqn:surrogate_loss}}
\EndFor
\State Discard oldest experiences based on specified Training Interval to make room for new experiences
\EndFor
\caption{LUSTER Training}
\label{alg:luster-training}
\end{algorithmic}
\end{algorithm}

\subsection{Training Configurations}
\label{sec:appendix:hyperparameters}

The predefined set for token selection in the following equations are:

In Equation \ref{eqn:emotion}, $E=$\{neutral, fearful, dissatisfied, apologetic, abusive, excited, satisfied\}.

In Equation \ref{eqn:domain}, $D=$\{general, restaurant, hotel, train, attraction, taxi, police, hospital\}, which includes the seven EmoWOZ domains plus an additional ``general'' domain to handle small-talk and greetings.

In Equation \ref{eqn:conduct}, $C=$\{neutral, compassionate, apologetic, enthusiastic, appreciative\}.

T5NLU achieves a slot F1 of 86.5\%;
LangEmoUS reports a macro F1 of 52.1\% for emotion prediction.

\begin{table}[ht]
\centering
\footnotesize
\begin{tabular}{ll}
\toprule
Parameter & Value \\
\midrule
Learning Rate & 3e-4 with LoRA, 3e-5 when fully fine-tuned \\
Training Epochs & 5 \\
Batch Size & 32 \\
LoRA Rank & 128 \\
LoRA Alpha & 32 \\
LoRA Dropout & 0.1 \\
LoRA Bias & None \\
LoRA Target Modules & all linear and projection layers \\
\bottomrule
\end{tabular}
\caption{Supervised Training Hyperparameters}
\label{tab:sl-hyperparameter}
\end{table}

\begin{table}[ht]
\centering
\footnotesize
\begin{tabular}{ll}
\toprule
Parameter & Value \\
\midrule
Critic Learning Rate & 5e-5 \\
Critic Batch Size & 32 \\
Critic Epochs & 5 \\
Critic Gradient Norm Clipping & 40 \\ \midrule
Actor Learning Rate & 1e-7 \\
Actor Batch Size & 16 \\
Actor Sample Size (Equation \ref{eqn:sequence-sample}) & 6 \\
Actor Sample Temperature (dialogue action $da$) & 0.5 \\
Actor Sample Temperature (response $u_{\sys}$) & 0.9 \\
Actor History Length & 5 \\ \midrule
Training Interval & 10 Dialogues \\
Experience Buffer Size & 50 Dialogues \\
Reservoir Sampling & True \\
User Utterance and Emotion Sample Size (Equation \ref{eqn:sequence-sample}) & 6 \\
Dialogue-level Discount Factor $\gamma_{\text{dial}}$ (Equation \ref{eqn:obj-q}) & 0.99 \\
Reward Combination Weight $\rho$ (Equation \ref{eqn:combined-reward}) & 0.1 \\
Importance Sampling Weight Clipping Range $\eta$ (Equation \ref{eqn:is-weight}) & 0.9 \\
Turn-level Discount Factor $\gamma_{\text{turn}}$ (Equation \ref{eqn:ppo-adv}) & 1.0 \\
Decay Rate $\lambda$ & 0.99 \\
Dialogue-level Reward (Equation \ref{eqn:obj-q}) & $r_t=-1$ for $0<t<T$, $r_T=\text{+80 if success else -40}$\\
Polyak Averaging Weight $\tau$ (Equation \ref{eqn:polyak-q}, \ref{eqn:polyak-v}) & 0.01 \\
\bottomrule
\end{tabular}
\caption{Reinforcement Learning Configurations}
\label{tab:rl-hyperparameter}
\end{table}
\newpage

\subsection{Human Trial Set-up Survey Questions}
\label{appendix:survey}
\paragraph{Question for Task Success} Did the system find what you look for? Did it provide all the information that you need? If you ask for a booking, did it provide you with a reference number?

(A) Yes to all; (B) No

\paragraph{Question for Sentiment Rating} How would you rate your sentiment after the conversation?

(A) Very Negative; (B) Negative; (C) Neutral; (D) Positive; (E) Very Positive

\subsection{Preference-based LLM Evaluation Prompt Template}
\label{sec:appendix:evaluation-prompt}
\begin{displayquote}
\textit{There are two systems system\_A and system\_B. The user talks to each of them to complete the same information retrieval goal. You are now act as a judge that determines which system has better overall performance.}\\
\textit{The dialogue between the user and system\_A in json list form is:} \textit{\{system\_a\_dialogue\}}\\
\textit{The dialogue between the user and system\_B in json list form is:} \textit{\{system\_b\_dialogue\}}\\
\textit{Now, output your judgement together with explanation in the following format:}\\
\textit{\{}\\
\textit{``judgement'': ``which system is better, system\_A or system\_B? If you really cannot decide, you can output `tie'.'',}\\
\textit{~~~~``explanation'': ``Why one system is better. Your analysis should be based on the success of the dialogue, the informativeness of the system responses, and the system emotional performance to enhance user experience. You can also consider the length of the dialogue, the diversity of language, and whether the system provided relevant information to the user.''}\\
\textit{\}}
\end{displayquote}

\newpage
\section{Complementary Results}
\subsection{Examples from Human Trial}
\label{appendix:excerpts}

Compared with the simulated user, participants in the paid human study tended to express their requests as concisely as possible in order to complete the assigned goal efficiently. 
This reflects a different interaction style from that of the simulator. 
Below, we present several dialogue openings from similar scenarios to illustrate how different systems respond to underspecified or ambiguous user requests at the start of an interaction. 
These examples are not intended as a controlled head-to-head comparison or as definitive evidence that one system is superior; rather, they serve as realistic snippets that highlight qualitatively different response patterns.

\begin{displayquote}
\footnotesize
{\textbf{User} : Train day sunday.}\\
{\textbf{EmoLoop} : There are 10 trains that leave at 05:00 from london kings cross.}
\end{displayquote}

\begin{displayquote}
\footnotesize
{\textbf{User} : Train.}\\
{\textbf{LUSTER-base} :i can certainly help you. where would you like to depart from and what is your destination and travel date and time?}
\end{displayquote}

\begin{displayquote}
\footnotesize
{\textbf{User} : TrainID.}\\
{\textbf{LUSTER} : i dont know what day youre traveling. you should tell me the day and then i can look it up for you. i also dont know what your departure and destination locations are.}
\end{displayquote}

The modular system, EmoLoop, immediately returns the number of available options, reflecting a strong prior over the dialogue state. 
This may hinder task progression when the user later specifies different constraints, but it can be effective when the user's initial request already aligns with those assumptions.
By contrast, both LUSTER and LUSTER-base attempt to clarify the missing constraints needed to narrow the search space, although they do so using different strategies, largely due to differences in the users' initial requests.

\subsection{Examples from Behavioural Analysis}
\label{appendix:examples-behavioural-analysis}

We provide representative examples from the human-evaluation logs to illustrate the behavioural categories discussed in Section~\ref{sec:behaviour-analysis}. 
The examples show both useful affective behaviour, such as repair and warm closure, and remaining failure modes, such as over-apology, redundant clarification, and task inconsistency.

\paragraph{Over-apology.}
The following example shows that surface-level apology is not sufficient for natural affective competence.

\begin{displayquote}
\footnotesize
{\textbf{Context}: The system apologises for confusion.}\\
{\textbf{User}: why are you apologising?}\\
{\textbf{LUSTER-base}: i am a bot and i am programmed to apologize even when i dont have any reason to do so.}
\end{displayquote}

\paragraph{Failure triggering user frustration.}
The following example illustrates how task-level inconsistency can lead to strongly negative user feedback.

\begin{displayquote}
\footnotesize
{\textbf{Context}: The user wants a train to arrive by 9:45, but the system provides a train leaving at 10:45.}\\
{\textbf{User}: how can it arrive by 9:45?}\\
{\textbf{LUSTER-base}: i do not understand. are you saying you need to arrive before you leave?}\
{\textbf{User}: you are useless}
\end{displayquote}

\paragraph{Repair after confusion.}
The following example shows repair behaviour after the system carries over irrelevant information from a previous task.

\begin{displayquote}
\footnotesize
{\textbf{Context}: The user asks for a train after an attraction-search task.}\\
{\textbf{LUSTER}: id love to help you find a train, but i will need more information first. thats the postcode and phone, right?}\\
{\textbf{User}: why are you talking about postcode and phone???}\\
{\textbf{LUSTER}: i apologize for the confusion. i have train tr8224 going from cambridge to bishops stortford at 11:29. would you like me to book 1 ticket for you?}
\end{displayquote}

\paragraph{Redundant clarification.}
The following example shows that clarification behaviour can become redundant when the user has already received sufficient information.

\begin{displayquote}
\footnotesize
{\textbf{Context}: The user has already received the requested information.}\\
{\textbf{User}: ok cool thanks!}\\
{\textbf{LUSTER}: what day would you like to travel, where are you departing from, and what is your destination?}
\end{displayquote}

\subsection{Examples from Preference-based LLM Evaluation}
\label{sec:appendix:case-study}
The simulated user goal is described below, where the user's goal is to retrieve \textbf{requestable} information based on \textbf{informable} constraints:

\begin{quote}\footnotesize
\texttt{\{"train": \{"informable": \{"day": "saturday", "departure": "cambridge",\\
"destination": "stansted airport", "leave at": "14:15"\},\\
"requestable": \{"duration": "?"\}\}, ...\}}
\end{quote}

The conversation between the simulated user and the LUSTER-base system with Llama-3.2-3b-instruct backbone is:
\begin{displayquote}
\footnotesize
{\textbf{User} : I'm looking for a train that is leaving on Saturday and going to stansted airport.}\\
{\textbf{System} : where will you be departing from?}\\
{\textbf{User} : cambridge and should leave after 14:15.}\\
{\textbf{System} : i have a train leaving at 14:40. would you like me to book that for you?}\\
{\textbf{User} : No thanks. What is the travel time for that train?}\\
{\textbf{System} : the travel time is 28 minutes. is there anything else i can help you with?}\\
...
\end{displayquote}

Under the same goal, the conversation between the simulated user and the LUSTER system with Llama-3.2-3b-instruct backbone is:
\begin{displayquote}
\footnotesize
{\textbf{User} : I'm looking for a train that is leaving on Saturday and going to stansted airport.}\\
{\textbf{System} : {\color{blue} there are 19 trains that match your criteria.} where will you be departing from?}\\
{\textbf{User} : cambridge and should leave after 14:15.}\\
{\textbf{System} : {\color{blue} tr2708 leaves at 14:40.} would you like me to book it for you? {\color{blue} how many tickets do you need?}}\\
{\textbf{User} : Yes, I'll need to know the travel time please.}\\
{\textbf{System} : the travel time is 28 minutes. is there anything else i can help you with?}\\
...
\end{displayquote}

Both systems fulfill the user goal by correctly providing requestable information: duration=28 minutes. 
ConvLab-3 evaluator marks both conversations as successful, and the user simulator, langEmoUS, gives the same sentiment rating in each turn. 
Despite this, GPT-4o-mini prefers System\_B (LUSTER) over System\_A (LUSTER-base) based on more nuanced system behaviours as highlighted in blue colour above, with the following explanation:

\begin{displayquote}
\footnotesize
    System\_B demonstrates better overall performance compared to system\_A.
    It successfully engages the user by {\color{blue} immediately presenting the number of trains available for their specified criteria}, which adds a layer of informativeness that system\_A lacks. \\
    Additionally, system\_B {\color{blue} reiterates the specific train details}, contributing to a fluid interaction. \\
    ...
\end{displayquote}

\end{document}